\documentclass{article}

\usepackage{cite}
\usepackage{amsmath,amssymb,amsfonts}
\usepackage{algorithmic}
\usepackage{graphicx}
\usepackage{textcomp}
\usepackage{amsmath,amssymb,amsfonts}
\usepackage{xcolor}
\usepackage{multirow}
\usepackage{rotating}
\usepackage{subcaption}
\usepackage{hyperref}
\usepackage{booktabs}
\usepackage{fancyhdr}
\usepackage{float}
\usepackage{threeparttable}
\usepackage{makecell} 

\usepackage[numbers]{natbib}

\usepackage{arxiv}

\usepackage[utf8]{inputenc} % allow utf-8 input
\usepackage[T1]{fontenc}    % use 8-bit T1 fonts
\usepackage{hyperref}       % hyperlinks
\usepackage{url}            % simple URL typesetting
\usepackage{booktabs}       % professional-quality tables
\usepackage{amsfonts}       % blackboard math symbols
\usepackage{nicefrac}       % compact symbols for 1/2, etc.
\usepackage{microtype}      % microtypography
\usepackage{lipsum}		% Can be removed after putting your text content
\usepackage{graphicx}
\usepackage{amsmath,amssymb,amsfonts}
\usepackage{natbib}
\usepackage{doi}
\usepackage{multicol}

\usepackage{threeparttable}
\usepackage{makecell}

\title{DB-FGA-Net: Dual Backbone Frequency Gated Attention Network for Multi-Class Brain Tumor Classification with Grad-CAM Interpretability}

\author{
Saraf Anzum Shreya\textsuperscript{1}\thanks{Email: sasshreya2001@gmail.com}, 
MD. Abu Ismail Siddique\textsuperscript{1}\thanks{Email: saif101303@gmail.com}, 
Sharaf Tasnim\textsuperscript{1}\thanks{Email: sharaftasnim786@gmail.com} \\
\textsuperscript{1}Dept. of Electronics and Telecommunication Engineering, \\Rajshahi University of Engineering and Technology, Rajshahi, Bangladesh
}

\renewcommand{\headeright}{}
\renewcommand{\undertitle}{}
\renewcommand{\shorttitle}{DB-FGA-Net: Dual Backbone Frequency Gated Attention Network for Multi-Class Brain Tumor Classification with Grad-CAM Interpretability}

\hypersetup{
pdftitle={A template for the arxiv style},
pdfsubject={q-bio.NC, q-bio.QM},
pdfauthor={David S.~Hippocampus, Elias D.~Striatum},
pdfkeywords={Currency Detection,  Computer Vision, Deep Learning, Machine Learning and YOLO},
}

\begin{document}
\maketitle

\begin{abstract}
	Brain tumors are a challenging problem in neuro-oncology, where early and precise diagnosis is important for successful treatment. Deep learning-based brain tumor classification methods often rely on heavy data augmentation which can limit generalization and trust in clinical applications. In this paper, we propose a double-backbone network integrating VGG16 and Xception with a Frequency-Gated Attention (FGA) Block to capture complementary local and global features. Our model achieves highly competitive performance without augmentation which demonstrates robustness to variably sized and distributed datasets. For further transparency, Grad-CAM is integrated to visualize the tumor regions based on which the model is giving prediction, bridging the gap between model prediction and clinical interpretability. The proposed framework achieves 99.24\% accuracy on the 7K-DS dataset for the 4-class setting, along with 98.68\% and 99.85\% in the 3-class and 2-class settings, respectively. On the independent 3K-DS dataset, the model generalizes with 95.77\% accuracy, outperforming several baseline methods under the same experimental setting. To further support clinical usability, we developed a graphical user interface (GUI) that provides real-time classification and Grad-CAM-based tumor localization. These findings suggest that augmentation-free, interpretable, and deployable deep learning models such as DB-FGA-Net hold strong potential for reliable clinical translation in brain tumor diagnosis.
\end{abstract}
% \begin{multicols}{2}
\maketitle

\section{Introduction}
\label{sec:introduction}
In general, a tumor is a clump of cells that grows out of control. Tumors can be classified into two categories: malignant and benign. Malignant tumors are cancerous and affects other parts of the body. On the other hand benign tumors grow slowly and are non-cancerous. When tumors occur in the brain or near it, is called a brain tumor. 
\\
Brain tumor is one of the worst tumors. It is ranked to be the 5th most common cancer by the ABTA organization\cite{CohenGadol2023BrainTumorStats}. It causes harm to the brain causing headaches to sensory loss, motor deficits even cognitive impairment \cite{Mukand2001BrainTumorRehab}. There are three common types of brain tumor; such as meningioma, glioma and pituitary. Tumors that arise from the meninges are meningiomas. Meninges are the membranes covering the brain and spinal cord. Meningiomas are mostly benign but can be malignant sometime. Glioma tumor occurs in the glial cells which is usually a malignant type of tumor. Pituitary tumor develops in the pituitary gland which is situated in the center of the brain. Pituitary tumors are generally benign.\cite{MayoClinic2024BrainTumor}
\\
According to the American Cancer Society, in 2023
approximately 25 thousand malignant brain and spinal cord tumors
will be diagnosed in the United States where 14,420 cases
in males and 10,980 in females \cite{ACS2025BrainStats}. More than 12,000 cases of primary brain tumors are recorded annually, including 500 children and young people which equates to 33 people every day \cite{BrainTumourCharity2025Statistics}. Early detection of brain tumors has significantly increased the survival rate as well as reduced the need for exploratory surgery to make a diagnosis. MRIs, CT scans, ultrasounds and PET scans are usually used to diagnose brain tumors \cite{Stieg2016}. Among them MRI provides better soft tissue contrast making MRI the most popular for brain tumor diagnosis, treatment planning and surgical guidance. \cite{MDAnderson2024}. Despite its advantages, manual diagnosis of brain tumor using MRI scans is labor-intensive. It is not immune to human error either. 
\\
Widely available CT imaging is useful in emergencies but it suffers from lower soft tissue contrast and exposes patients to ionizing radiation \cite{MDAnderson2024}. Although PET is effective in measuring tumor metabolism it has limited accessibility and on top of that it is expensive. Furthermore, traditional diagnosis heavily rely on invasive biopsy for definitive diagnosis of brain tumor, which is prone to infection, bleeding and complications\cite{MayoClinic2024BrainTumor}. These limitations beg for the urgent need for robust, automated, and non-invasive diagnostic solutions.

\subsection{Contributions of This Work}

In this work, a novel architecture that integrates VGG16 and Xception with a newly designed FGA block, effectively combining fine-grained local details with global contextual features to improve tumor classification. Unlike most prior approaches that rely heavily on data augmentation, our model achieves strong performance without relying on data augmentation, reducing preprocessing requirements and emphasizing its practical usability in real-world clinical settings. The performance of the proposed model is further validated on an independent dataset (3K-DS), demonstrating its robustness across different data distributions.
\\

A comparative evaluation was conducted against CBAM based baselines, demonstrating improved or competitive performance and enhanced interpretability compared to CBAM baselines through quantitative metrics and Grad-CAM visualizations. To ensure clinical trust, we leverage explainable AI techniques to visualize the model’s predictions, showing that decisions are based on tumor regions rather than irrelevant areas of the images. Furthermore, we develop a lightweight graphical user interface (GUI) that integrates both classification and Grad-CAM visualizations, allowing clinicians to interactively validate tumor predictions and their corresponding locations. This combination of accuracy, interpretability, and usability underscores the practical contributions of our work.

\section{Related Works}

\begin{table*}[htbp]
\centering
\caption{Comprehensive Comparison of Brain Tumor Classification Models}
% Adjust table spacing
\renewcommand{\arraystretch}{1.5} % row height
\setlength{\tabcolsep}{6pt} % column padding
\label{tab:model_comparison}
\resizebox{\textwidth}{!}{%
\begin{tabular}{|p{1.5cm}|p{1.5cm}|p{1.5cm}|p{2.5cm}|p{3cm}|p{2cm}|p{1.2cm}|p{1.2cm}|p{1.2cm}|}
\hline
\textbf{Authors} & \textbf{Publication Date} & \textbf{Dataset} & \textbf{Augmentation} & \textbf{Model} & \textbf{Accuracy} & \textbf{Precision} & \textbf{Recall} & \textbf{F1-Score} \\ \hline
P. Chauhan et al.\cite{PBviT} & 23 Dec 2024 & Figshare (3 class) & Rotation (90), Shear, Saturation Adjustment & Patch-Based Vision Transformer & 95.8\% & 95.3\% & 93.2\% & 92\% \\ \hline
A. Saeed et al. \cite{GGLA} & 3 Jan 2025 & 7K-DS, Figshare (3 class) & GAN / Simple augmentation & Dual-Branch Gated Attention Network & 96.87--99.62\% & 96.42--99.06\% & 96.32--99.05\% & 96.73--99.62\% \\ \hline
N. Sivakumar et al. \cite{FedAvg} & 10 Mar 2025 & 7K-DS & N/A & CNN + FedAvg + FedProx & 97.19\% & High & High & 97.18\% \\ \hline
Anees Tariq et al. \cite{trans} & 28 Mar 2025 & 7K-DS & Rotation, flipping, scaling, cropping, color adjustments & ViT + EfficientNetV2 (Ensemble) & 96\% & 96\% & 96\% & 96\% \\ \hline
R. Preetha et al. \cite{hybrid3b} & 7 Apr 2025 & Combined 4 datasets & Rotation, scaling, flipping, mirroring, cropping & Hybrid 3B Net + EfficientNetB2 & 97.80\% (4-class), 98.72\% (3-class), 99.50\% (2-class) & High & High & High \\ \hline
N. M. Hussain Hassan et al. \cite{fuzzy} & 1 May 2025 & 7K-DS, 3K-DS, 13K-DS & Rotation (0,90,180,270) & Fuzzy Thresholding + DL & 98.42\%--99.42\% & 98.16\%--98.65\% & 98.14\%--98.26\% & 98.1\%--98.65\% \\ \hline
H. Alshaari et al. \cite{deep} & 7 May 2025 & 7K-DS, 3K-DS & N/A & EfficientNetB0 w/ Dual Reg. & 98\% & 95\% & 98.2\% & 95.4\% \\ \hline
R. D. Prayogo et al. \cite{hybrid-cnn} & 30 Jun 2025 & 7K-DS & Rotation, horizontal flip & ResNet50V2 + MobileNetV2 + DenseNet121 & 98.75\% & 98.76\% & 98.75\% & 98.75\% \\ \hline
\end{tabular}%
}
\end{table*}

The paper \cite{hybrid3b} proposed a three-branch convolutional neural network (3B Net) integrated with EfficientNetB2 for brain tumor classification. They also compared their proposed model with other multiple-branch models, branches ranging from 1 to 6. Each branch extracts distinct feature sets from the input MRI images, which are then fused using a concatenation layer followed by fully connected layers for classification. The model was trained on both augmented and non-augmented datasets and it was also trained on both binary(tumor vs. no-tumor) and multi-class dataset. For 4-class dataset, the augmented dataset had a higher accuracy of 98.70\% while the non-augmented dataset had 94.95\%. Authors of the paper \cite{hybrid-cnn} proposed hybrid models that concatenated features from the best performing transfer learning models. Features are extracted from the last three blocks of each model, then concatenated and passed through a global average pooling layer before a softmax classifier. Augmentation parameters include a shear range of 0.2 and rotation range of 30 degrees. Their approach achieved accuracy of 98.42\% to 98.75\%. A dual-branch ensemble with Gated Global-Local Attention (GGLA) \cite{GGLA} uses EfficientNetV2S and ConvNeXt, dataset enhanced by ESRGAN for data balancing and preprocessing for noise reduction. It achieves 99.06\% and 99.62\% accuracy on three and four-class datasets. Their proposed model shows good localization of the tumor area via Grad-CAM analysis.
\\
The paper \cite{content} proposes a hybrid model that uses pre-trained EfficientNet-B0 and Local Binary Pattern (LBP) for feature extraction, with minimum redundancy maximum relevance (mRMR) to prune redundant features. It classifies three diseases (Alzheimer’s, multiple sclerosis, intracranial regions) and eight classes, achieving 98.9\% accuracy on a diverse dataset. A fine tuned  EfficientNetB0 model was suggested by the authors H. Alshaari et. al \cite{deep}. Evaluated on 3,064 and 7,023-image datasets, it achieves 95\% and 98\% accuracy respectively. 
\\
A research \cite{PBviT} presents PBVit, a patch-based Vision Transformer (ViT) specifically designed to improve brain tumor detection using MRI images. Their methodology suggests dividing each MRI image into fixed-size patches, typically 16$\times$16 pixels. These patches are treated as tokens and linearly projected into lower-dimensional embeddings. Those dimensions with positional encodings added to retain spatial relationships. These tokens are processed through multiple transformer layers, each consisting of multi-head self-attention mechanisms and feed-forward networks. The model is evaluated and got an accuracy of 95.8\%, precision of 95.3\%, recall of 93.2\%, and an F1-score of 92\%.
\\
The authors Chaki et. al \cite{retrival} introduces the Deep Brain IncepRes Architecture 2.0 based Reinforcement Learning Network (DBIRA2.0-RLN). This CNN architecture was designed for brain tumor classification and retrieval from MRI images. The architecture incorporates Inception blocks that extract multi-scale features using 1$\times$1, 3$\times$3, and 5$\times$5 convolutions, complemented by skip connections to mitigate vanishing gradient issues and improve training stability. Fuzzy logic enhances the approach by applying membership functions to weight the importance of different features, while Multilinear Principal Component Analysis (MPCA) reduces dimensionality by retaining 90\% of the variance in the descriptor vectors. The model classifies tumors into meningioma, glioma and pituitary types; and detects non-tumor cases, achieving accuracies of 97.1\%, 98.7\%, 94.3\%, and 100\% respectively. This study \cite{trans} proposes a hybrid approach that combines EfficientNetV2 and Vision Transformer (ViT) models to enhance multi-class brain tumor classification. EfficientNetV2 extracts hierarchical features using its compound scaling technique which adjusts width, depth, and resolution, while ViT processes image patches with self-attention mechanisms to capture global dependencies. Both models were evaluated on 7,023 image MRI dataset. A geometric mean ensemble learning technique fuses the predictions from EfficientNetV2 and ViT, weighting them based on validation set performance. It achieved accuracies of 95\% (EfficientNetV2), 90\% (ViT) and 96\% (ensemble) for multi-class tasks.
\\
N. Sivakumar et. al \cite{FedAvg} introduce a hybrid brain tumor classification method using federated learning (FL) with FedAvg and FedProx algorithms to train CNNs across decentralized datasets. The methodology suggests a base CNN architecture, featured with three convolutional layers and two dense layers, across five client devices, each holding a subset of the Kaggle MRI dataset to preserve privacy while training. FedAvg aggregates model updates from clients using a weighted average based on data size, while FedProx adds a proximal term to handle data heterogeneity, improving convergence stability. Another study \cite{abnormal} proposes an enhanced ResNet50 model for classifying abnormal brain tumors using MRI images to address diagnostic challenges and improve precision in early detection. The methodology leverages transfer learning by fine-tuning ResNet50 with ImageNet weights, incorporating data augmentation techniques such as random rotations, flips and zooms to increase dataset diversity and prevent overfitting. The model is trained on a dataset of 5712 MRI images, achieving 99\% accuracy in both training and validation phases.
\\
Attention mechanisms have become a cornerstone in deep learning, particularly for enhancing the performance of convolutional neural networks (CNNs) by focusing on the most relevant features within an input. These mechanisms allow models to weigh the importance of different regions or channels dynamically, improving both accuracy and interpretability. The seminal work by Vaswani et al. \cite{vaswani2017attention} introduced the Transformer architecture, which popularized self-attention for natural language processing and later influenced computer vision tasks. In the context of image classification, attention mechanisms help mitigate the limitations of CNNs by selectively emphasizing informative features while suppressing irrelevant ones, a principle that has been widely adopted in medical imaging to improve diagnostic precision.
\\
Channel attention and spatial attention are two fundamental variants that have been extensively explored to refine feature representations. The Squeeze-and-Excitation Network (SE-Net) by Hu et al. \cite{hu2018squeeze} pioneered channel attention by recalibrating channel-wise feature responses through global average pooling and fully connected layers, achieving significant improvements on ImageNet (e.g., 1\% top-1 accuracy gain). This approach inspired subsequent works to focus on inter-channel relationships, enhancing feature discriminability. Spatial attention, on the other hand, emphasizes where to look within an image. Wang et al. \cite{wang2017residual} proposed the Residual Attention Network, which integrates spatial attention to refine feature maps by learning to focus on salient regions, demonstrating enhanced performance on object recognition tasks. The Convolutional Block Attention Module (CBAM) by Woo et al. \cite{woo2018cbam} further combined both channel and spatial attention sequentially, improving classification and detection tasks by up to 1.2\% on MS COCO and VOC datasets. These dual-attention mechanisms have been adapted for medical imaging, with studies like \cite{schlemper2019attention} applying them to MRI segmentation to highlight tumor boundaries, underscoring their relevance for precise feature extraction.
\\
Beyond spatial and channel attention, frequency domain attention has gained traction as a complementary approach to capture periodic patterns and spectral characteristics, particularly valuable for texture-rich medical data such as MRI scans. However, CBAM's focus on spatial/channel domains overlooks spectral characteristics in frequency-rich medical data, such as tumor textures in brain MRIs, which our proposed Frequency-Gated Attention (FGA) addresses through Fourier-based gating. 
\\
Xu et al. \cite{xu2020frequency} introduced the Frequency Attention Network (FAN) for image classification, leveraging a frequency attention module in the Fourier domain to modulate features by emphasizing high-frequency components (e.g., edges) and low-frequency ones (e.g., global structure). This method improved ImageNet classification accuracy by 1.5\% over baseline ResNet models, highlighting the potential of spectral analysis. Similarly, Li et al. \cite{li2021frequency} developed a frequency domain attention module for medical image segmentation, using Fast Fourier Transform (FFT) to extract frequency features and a gating mechanism to fuse them with spatial attention. This approach achieved a 2.3\% Dice score improvement on the BraTS dataset for brain tumor segmentation, demonstrating its efficacy in capturing texture details critical for tumor delineation. These advancements motivate the integration of frequency attention in our proposed Frequency Gated Attention (FGA) block, aiming to enhance feature representation by incorporating spectral information without significant computational overhead.
\\
Despite notable progress in brain tumor classification, several limitations remain in existing literature. Many existing methods rely heavily on data augmentation or synthetic data generation using GANs, which may not fully capture the natural variability of medical imaging. While these approaches often achieve high reported accuracy, their performance on unseen datasets is rarely evaluated, raising concerns about generalization and clinical applicability. Furthermore, interpretability is frequently overlooked, with most models operating as black boxes without providing sufficient insight into tumor localization. Another limitation is that several studies restrict their evaluations to binary or three-class settings, which simplifies the diagnostic challenge but reduces the relevance of such models for real-world, four-class clinical tasks.

In this work, we address these limitations by proposing a Frequency-Gated Attention (FGA) enhanced dual-backbone framework. Unlike prior methods, our approach achieves high performance without reliance on augmentation, thereby reducing dependency on artificially inflated datasets. To ensure robustness, we perform cross-dataset validation, confirming the ability of the model to generalize to independent MRI collections. Interpretability is incorporated through Grad-CAM visualizations, which highlight tumor-relevant regions and provide transparency in decision-making. Finally, our framework is evaluated on binary, three-class, and four-class configurations, ensuring its clinical relevance across varying diagnostic scenarios. Together, these contributions establish DB-FGA-Net as a reliable and interpretable solution for brain tumor classification.

\section{Dataset Descriptions}
To conduct the training of the model a dataset was chosen from the kaggle which has a total of 7,023 images. We will be referring to it as 7K-DS and another dataset that was used has 3,264 images which will be referred to as 3K-DS.

\begin{table}[h]
\caption{Detailed dataset description of our primary dataset 7K-DS}
% Adjust table spacing
\renewcommand{\arraystretch}{1.5} % row height
\setlength{\tabcolsep}{6pt} % column padding
\centering
\begin{tabular}{|l|c|c|}
\hline
\textbf{Category} & \textbf{Training Set} & \textbf{Testing Set} \\ \hline
Glioma            & 1321                 & 300                  \\ \hline
Meningioma        & 1339                 & 306                  \\ \hline
Notumor           & 1595                 & 405                  \\ \hline
Pituitary         & 1457                 & 300                  \\ \hline
\textbf{Total}    & \textbf{5712}        & \textbf{1311}        \\ \hline
\end{tabular}
\label{tab:primary_dataset}
\end{table}

\begin{table}[h]
\caption{Detailed dataset description of our secondary dataset 3K-DS}
% Adjust table spacing
\renewcommand{\arraystretch}{1.5} % row height
\setlength{\tabcolsep}{6pt} % column padding
\centering
\begin{tabular}{|l|c|}
\hline
\textbf{Category} & \textbf{Images}  \\ \hline
Glioma            & 926                                   \\ \hline
Meningioma        & 937                                  \\ \hline
No Tumor          & 500                                   \\ \hline
Pituitary         & 901                                    \\ \hline
\textbf{Total}    & \textbf{3264}              \\ \hline
\end{tabular}
\label{tab:secondary_dataset}
\end{table}

\subsection{Primary Dataset (7K-DS)}
The 7K-DS dataset was collected from kaggle \cite{7k-ds} and IEEE DataPort \cite{7k}, which is widely adopted in prior works. It is a combination of 3 public datasets; figshare \cite{figshare}, SARTAJ dataset \cite{sartaj} and Br35H \cite{Br35H}. This dataset contains a total of 7,023 images of human brain MRI. Among them 5,712 images were for training and 1,311 images for testing. It has 4 classes, 3 tumor classes and a no tumor class. The dataset has already been divided into training and testing subsets. We are using this dataset as our primary dataset for this work. We have trained and tested our model on this. We have made two additional datasets from this dataset where one contains only 3 tumor classes and the other one has 2 classes (tumor and no-tumor class). Table~\ref{tab:primary_dataset} shows the details of the 7K-DS dataset.

\begin{figure}[htbp]
    \centering
    \begin{subfigure}[b]{0.4\linewidth}
        \includegraphics[width=\linewidth]{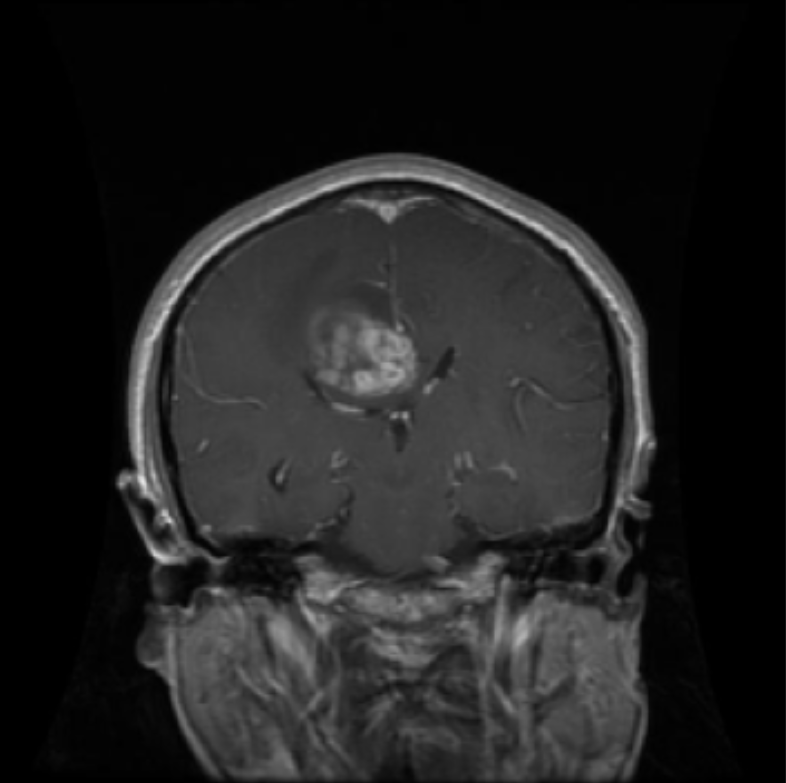}
        \caption{Glioma}
    \end{subfigure}
    \hfill
    \begin{subfigure}[b]{0.4\linewidth}
        \includegraphics[width=\linewidth]{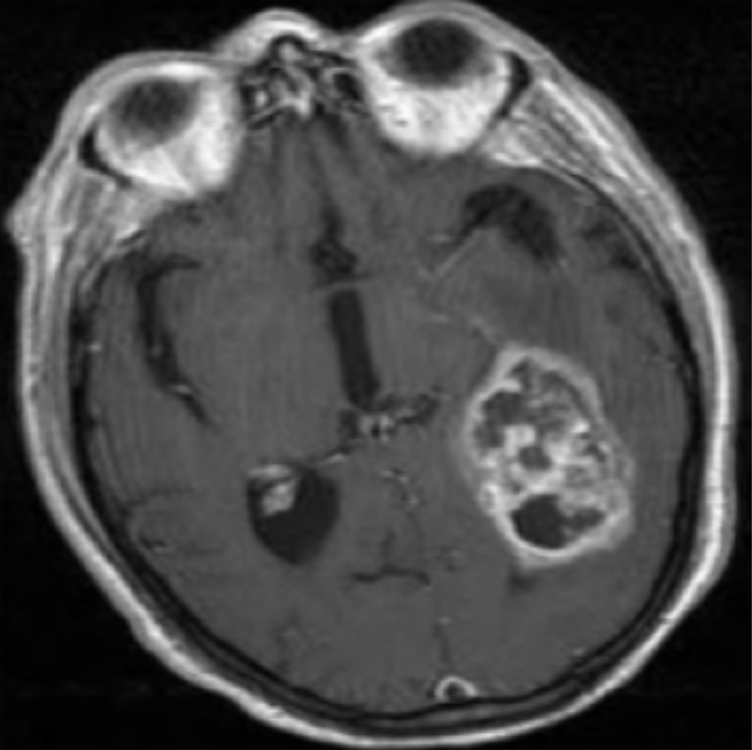}
        \caption{Meningioma}
    \end{subfigure}\\
    % \hfill
    \begin{subfigure}[b]{0.4\linewidth}
        \includegraphics[width=\linewidth]{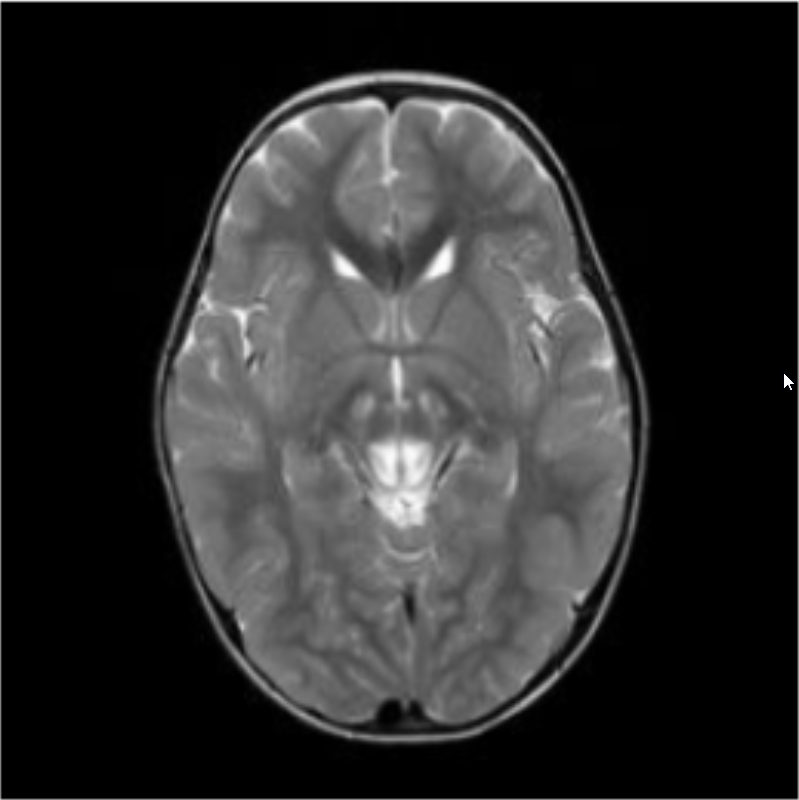}
        \caption{No-tumor}
    \end{subfigure}
    \hfill
    \begin{subfigure}[b]{0.4\linewidth}
        \includegraphics[width=\linewidth]{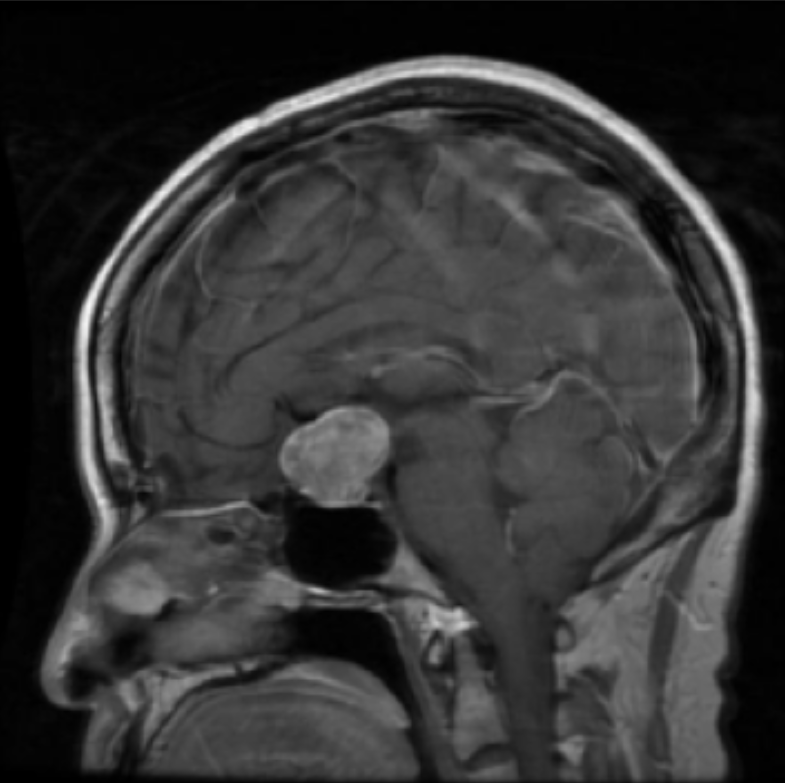}
        \caption{Pituitary}
    \end{subfigure}
    
    \caption{Sample MRI images from the 7K-DS dataset 
    (Glioma, Meningioma, No Tumor and Pituitary).}
    \label{fig:dataset_samples}
\end{figure}

\begin{figure}[htbp]
    \centering
    \begin{subfigure}[b]{0.4\linewidth}
        \includegraphics[width=\linewidth]{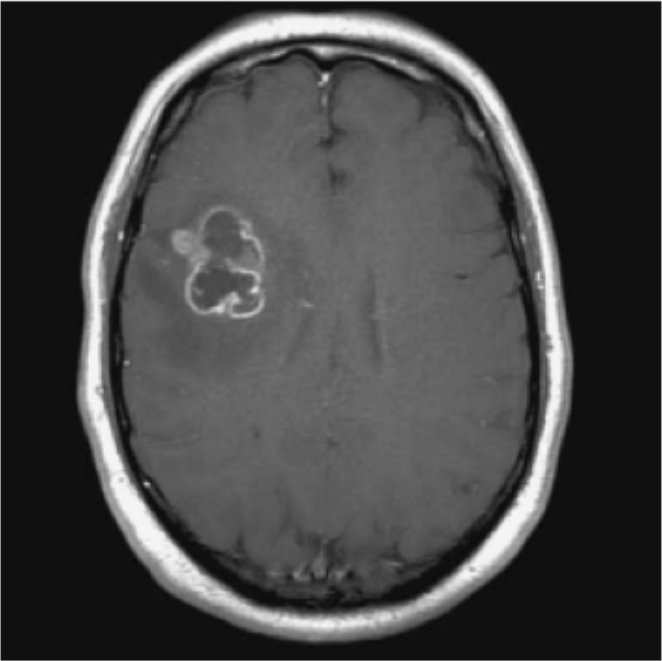}
        \caption{Glioma}
    \end{subfigure}
    \hfill
    \begin{subfigure}[b]{0.4\linewidth}
        \includegraphics[width=\linewidth]{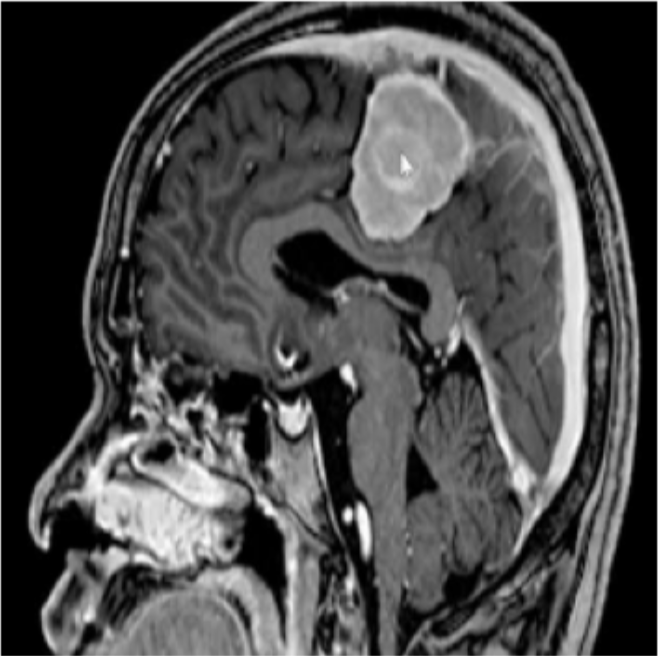}
        \caption{Meningioma}
    \end{subfigure}\\
    % \hfill
    \begin{subfigure}[b]{0.4\linewidth}
        \includegraphics[width=\linewidth]{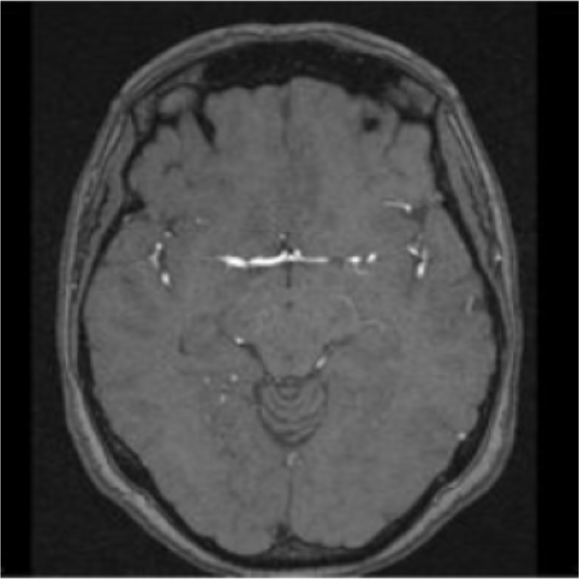}
        \caption{No-tumor}
    \end{subfigure}
    \hfill
    \begin{subfigure}[b]{0.4\linewidth}
        \includegraphics[width=\linewidth]{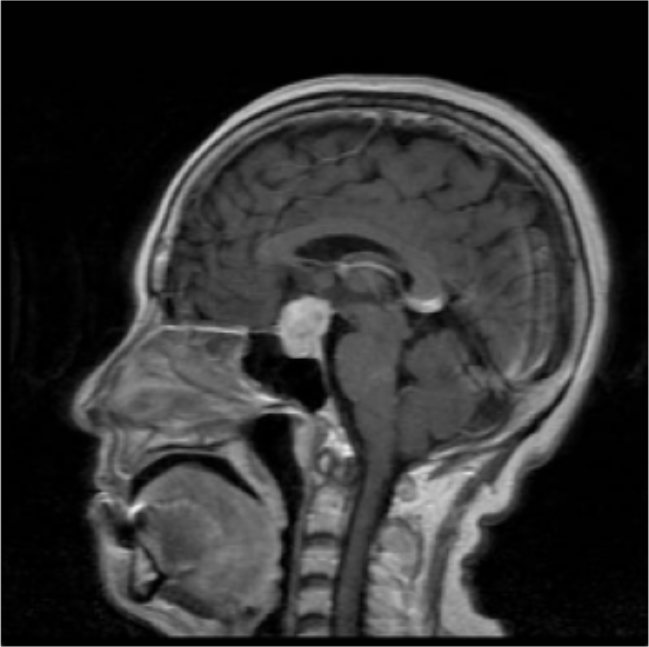}
        \caption{Pituitary}
    \end{subfigure}
    
    \caption{Sample MRI images from the 3K-DS dataset 
    (Glioma, Meningioma, No Tumor and Pituitary).}
    \label{fig:dataset_samples}
\end{figure}

\begin{figure*}[h]
    \centering
    \includegraphics[width=1\textwidth]{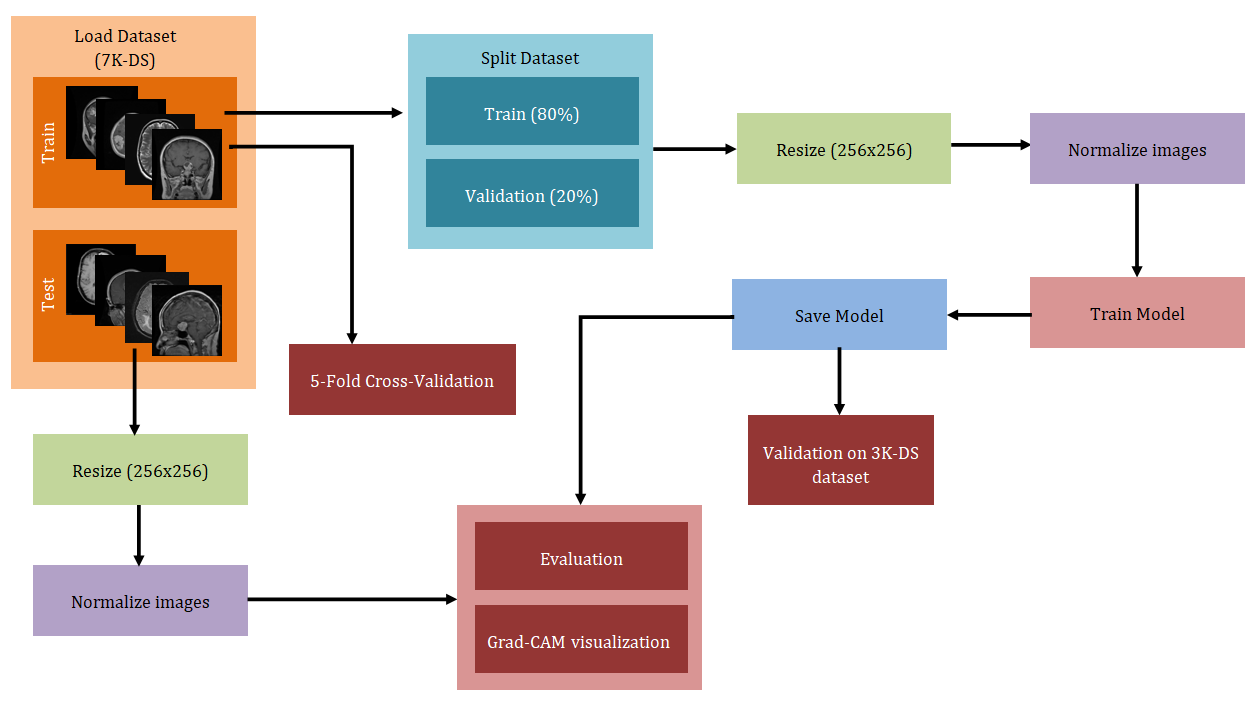}
    \caption{Visual representation of the Workflow of the proposed approach.}
    \label{fig:workflow}
\end{figure*}

\subsection{Secondary Dataset (3K-DS)}
The secondary dataset that was used was also collected from kaggle \cite{3k-ds}. A public dataset that contains 3,264 MRI images of brains. It also has 4 classes like the 7K-DS dataset. The whole dataset was used to check the validity of the proposed model. Figure~\ref{tab:secondary_dataset} shows the details of the secondary dataset.

\subsection{Preprocessing}
The datasets, 7K-DS and 3K-DS, were preprocessed to ensure compatibility with the deep learning models. Each image was resized to a uniform resolution of $256 \times 256$ pixels. No data augmentation techniques, such as rotation, flipping, or shearing, were applied as the proposed framework aims to demonstrate robustness without reliance on synthetic data.

Following resizing, each pixel value in the RGB image which originally was in the range $[0, 255]$, was converted to a floating-point format and scaled to the range $[0, 1]$. For an input image $I$ with pixel intensities $I(x, y, c)$ at spatial coordinates $(x, y)$ and color channel $c \in \{R, G, B\}$, the normalization is defined as:
\begin{equation}
I_{\text{norm}}(x, y, c) = \frac{I(x, y, c)}{255}
\end{equation}
where $I_{\text{norm}}(x, y, c)$ represents the normalized pixel value. This step ensures that the input data to the convolutional neural network (CNN) and FGA-enhanced architectures are within a consistent range.

Labels were processed by mapping class names to integer indices based on the sorted order of class folders. For a dataset with $C$ classes (e.g., $C = 4$ for 4-class, $C = 3$ for 3-class, or $C = 2$ for 2-class settings), each label $l_i \in \{0, 1, \ldots, C-1\}$ was converted to a one-hot encoded vector using TensorFlow's \texttt{to\_categorical} function. The one-hot encoding for a label $l_i$ is defined as:
\begin{equation}
y_i = [y_{i,0}, y_{i,1}, \ldots, y_{i,C-1}], 
\text{where} \quad y_{i,j} = \begin{cases} 
1 & \text{if } j = l_i, \\
0 & \text{otherwise}.
\end{cases}
\end{equation}
This encoding produces a $C$-dimensional vector for each sample.

For validation, the training set was split into training and validation set by 80/20. The preprocessing pipeline ensures that input images are uniformly formatted and normalized and labels are appropriately encoded for multi-class tumor classification.

\section{Proposed Methodology}

The overall workflow of the proposed brain tumor classification framework is illustrated in Figure~\ref{fig:workflow}. The aim is to develop a model that not only accurately classifies tumors but also localizes the tumor area. We have trained the model on a variety of classes; the 4-class dataset has three tumor classes and one no-tumor class, the 3-class dataset has only the three tumor classes and the 2-class dataset has tumor and no-tumor class. We trained and tested on the primary dataset. Before training the train subset was split in train and validation subsets and then resized and normalized. It was then trained on the datasets with a variety of class numbers. The trained models was tested on the secondary dataset to ensure the generality of the model. We also did 5 fold cross-validation to ensure that the result was not coincidental rather a solid outcome. The model was further evaluated on several metrics and most importantly, the Grad-CAM visualization was done to check it the model is able to make decisions on the actual tumor area.
\\
Now for the model, we first trained on a few base models such as VGG16, VGG19, MobileNetV2 and Xception. Then these same models with FGA (Frequency Gated Attention) block. The dual backbone FGA attention model has surpassed our other models in performance as well as in highlighting the tumor area. The model was deployed through a GUI built in Python using Tkinter, TensorFlow, OpenCV, and Pillow libraries, which provides an interactive platform for uploading MRI scans, viewing predictions, and observing Grad-CAM-based tumor localization.
\\

\subsection{Baseline Architectures}

\begin{figure}[h]
    \centering
    \includegraphics[width=0.5\textwidth]{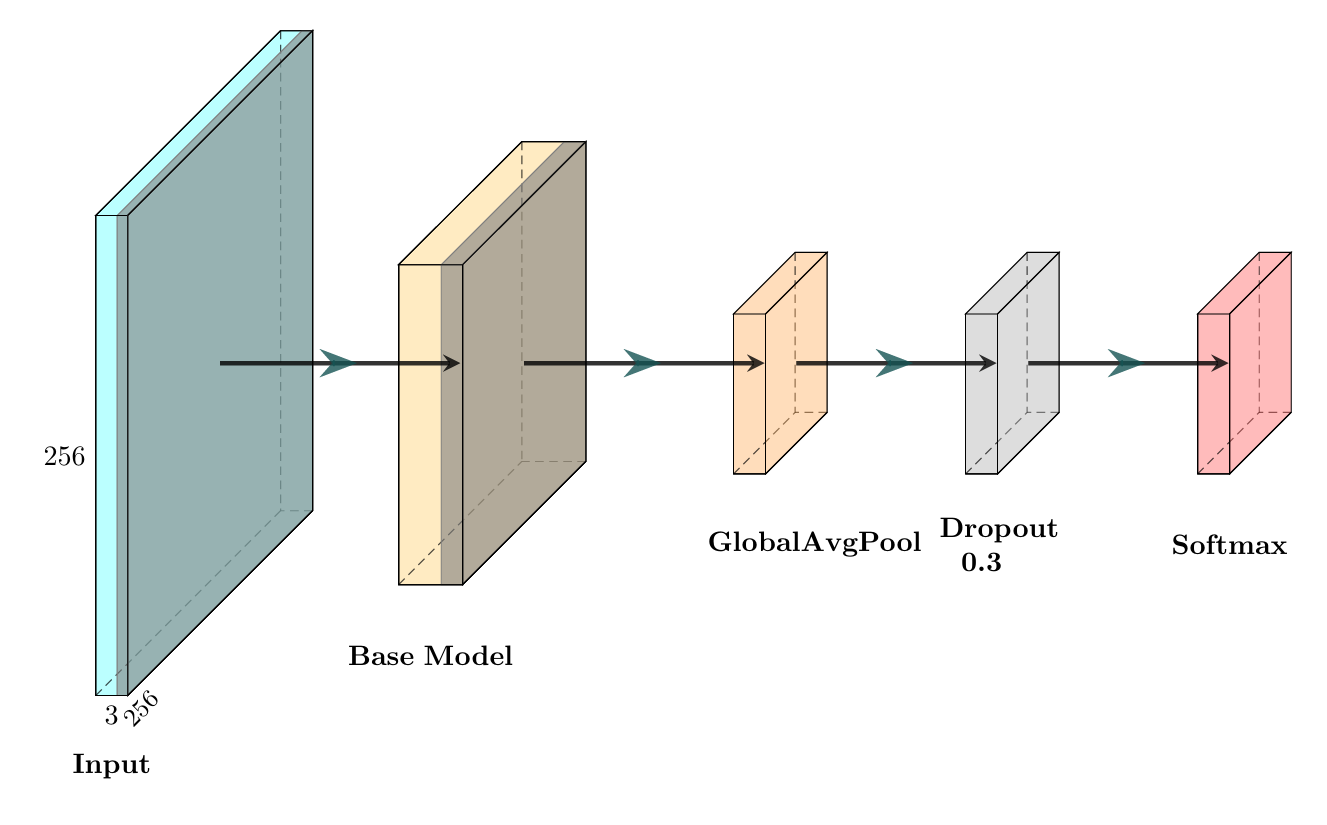}
    \caption{Baseline architecture: input images are processed through a CNN backbone, followed by Global Average Pooling, Dropout, and a Softmax classifier.}
    \label{fig:base_model}
\end{figure}

\begin{figure*}[h]
    \centering
    \includegraphics[width=1\textwidth]{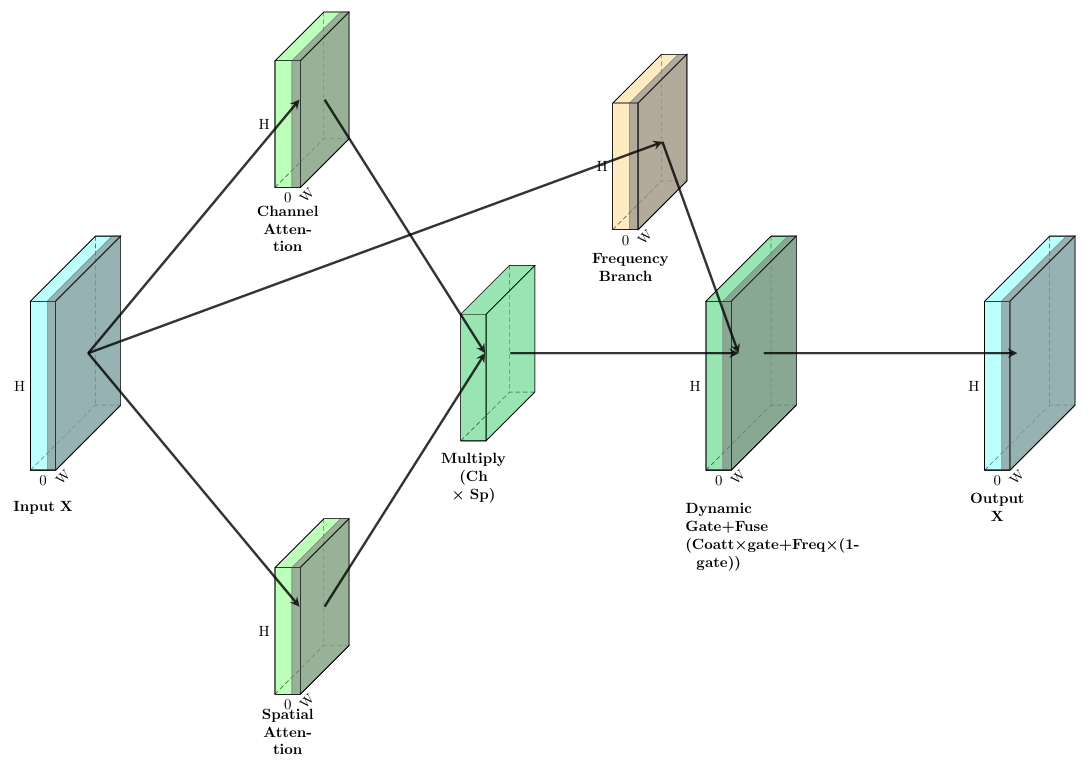}
    \caption{Detailed structure of the FGA block, illustrating the sequential channel and spatial attention mechanisms applied to enhance feature maps in the CNN architecture.}
    \label{fig:FGA_block}
\end{figure*}

\begin{figure*}[h]
    \centering
    \includegraphics[width=0.8\textwidth]{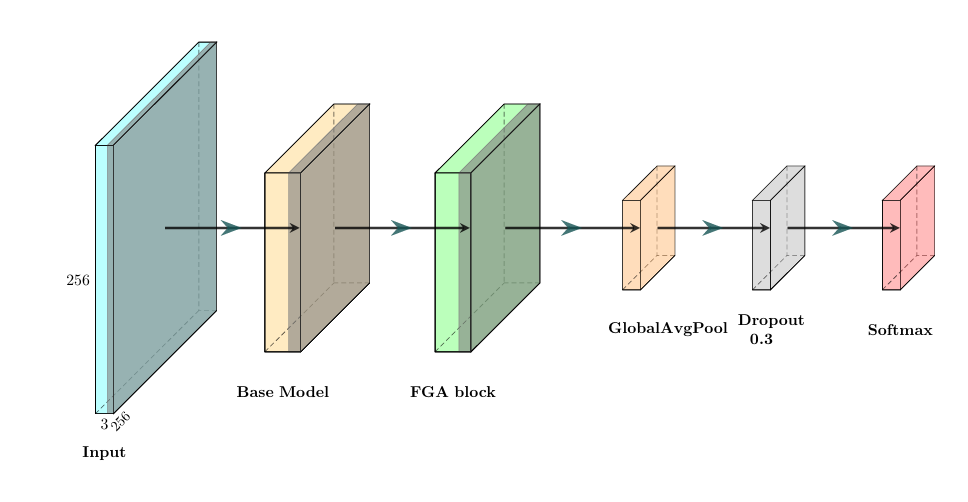}
    \caption{Integrated Base+FGA architecture, showing the CNN backbone enhanced with FGA, followed by Global Average Pooling, Dropout, and a Softmax classifier.}
    \label{fig:base_FGA}
\end{figure*}

\begin{figure*}[h]
    \centering
    \includegraphics[width=1\textwidth]{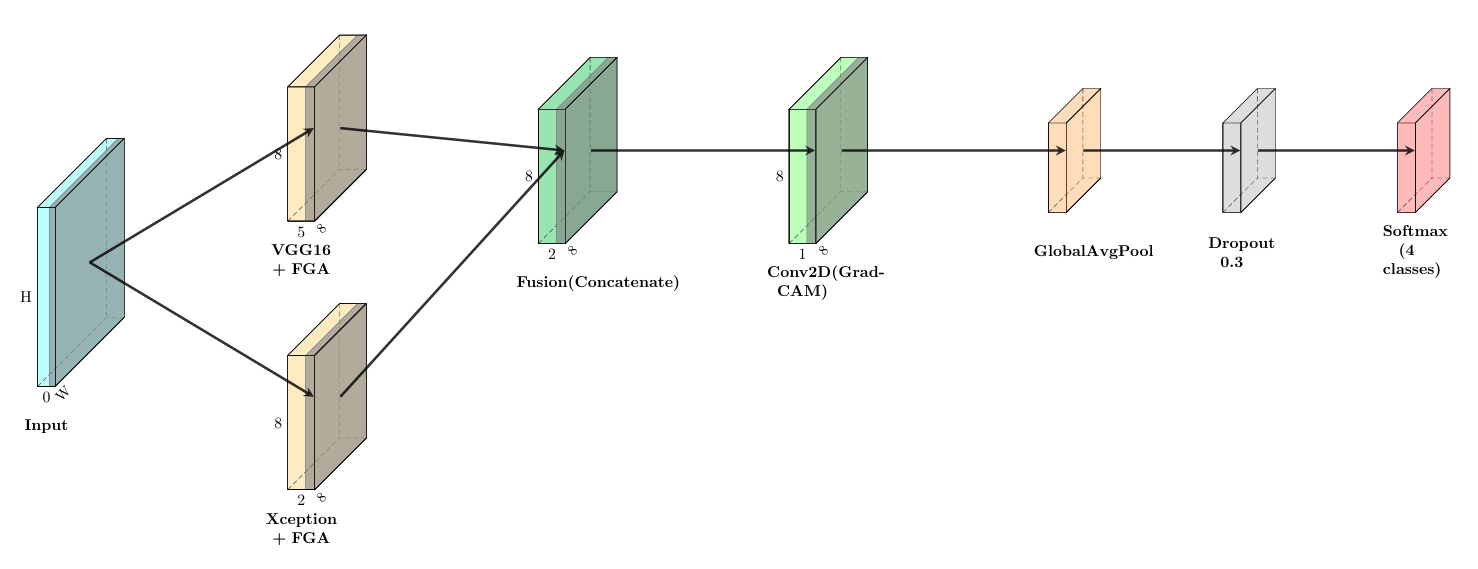}
    \caption{Dual-backbone architecture combining FGA-enhanced VGG16 and Xception.}
    \label{fig:dual_backbone}
\end{figure*}

The foundation of the proposed framework relies on a suite of pre-trained convolutional neural network (CNN) architectures such as VGG16, VGG19, MobileNetV2, and Xception. These models were selected for their established efficacy in feature extraction and adaptability to medical imaging tasks, particularly brain tumor classification from MRI scans. Pre-trained on the ImageNet dataset comprising over 1.2 million images across 1,000 classes, provide a robust starting point for transfer learning, leveraging low-level features (e.g., edges, textures) and high-level semantic information (e.g., object shapes) to identify tumor characteristics~\cite{simonyan2014very, sandler2018mobilenetv2, chollet2017xception}. Input images are resized to a uniform resolution of $256\times256\times3$ (RGB channels), preserving critical anatomical details while aligning with the models' input requirements. As depicted in Figure~\ref{fig:base_model}, the architectural pipeline begins with the CNN backbone, where convolutional layers extract spatial hierarchies of features. For VGG16 and VGG19, the architecture comprises 13 and 16 convolutional layers respectively, organized into five blocks with $3\times3$ filters and increasing channel depths (64 to 512), interspersed with max-pooling layers to reduce spatial dimensions, culminating in a feature map of $512 \times 8 \times 8$ after the final block. This depth enables fine-grained feature extraction but introduces higher computational complexity, with approximately 138 million parameters for VGG16~\cite{simonyan2014very}. In contrast, MobileNetV2 employs depthwise separable convolutions, splitting standard convolutions into a depthwise convolution (applying a single filter per input channel) and a $1 \times 1$ pointwise convolution, reducing parameter count to about 3.5 million while maintaining performance through inverted residuals and linear bottlenecks~\cite{sandler2018mobilenetv2}. Xception extends this efficiency with extreme inception modules, replacing traditional convolutions with depthwise separable convolutions across 36 layers, augmented by residual connections to mitigate vanishing gradients, producing a feature map of $2048\times4\times4$ with approximately 22 million parameters~\cite{chollet2017xception}. The output of each backbone is processed through Global Average Pooling (GlobalAvgPool), compressing the spatial dimensions into a single $1\times1\times C$ vector (where $C$ is the number of channels), followed by a dropout layer with a 0.3 rate to regularize the model by randomly deactivating 30\% of neurons during training, thus preventing overfitting. A final dense layer with softmax activation, tailored to the number of classes (4, 3, or 2 based on the classification task), produces probability distributions over tumor types (e.g., glioma, meningioma, no tumor, pituitary). The baseline models are trained without attention mechanisms, serving as a control to quantify the performance gains from FGA integration, with optimization performed using the Adam optimizer at a learning rate of $1 \times 10^{-4}$ and categorical cross-entropy loss, consistent with the multi-class nature of the 7K-DS dataset~\cite{kingma2014adam}.

%%%%%%%%%%%%%%%%%%%%%%%%%%%%%%%%%%%%

\subsection{FGA Block Integration}

The Frequency Gated Attention (FGA) block is a vital component of our work which is designed to enhance feature extraction by integrating channel-spatial co-attention and frequency domain attention with dynamic gating. The detailed architecture of the FGA block is depicted in Figure~\ref{fig:FGA_block}, which illustrates the sequential and fused attention mechanisms applied to the input feature maps. The integration of the FGA block into the base CNN backbone, followed by global pooling, dropout, and classification layers, is shown in Figure~\ref{fig:base_FGA}. Additionally, the dual-backbone architecture combining FGA-enhanced VGG16 and Xception is presented in Figure~\ref{fig:dual_backbone}.
\\
The FGA block operates on an input feature map \( X \in \mathbb{R}^{H \times W \times C} \), where \( H \), \( W \), and \( C \) denote height, width, and channels, respectively. The block consists of three main sub-modules: channel-spatial co-attention, frequency attention, and dynamic gating, fused with a residual connection.
\\
\subsubsection*{Channel-Spatial Co-Attention}
The channel attention mechanism begins with global average pooling to generate a channel descriptor \( F_c \in \mathbb{R}^{1 \times 1 \times C} \), computed as:
\begin{equation}
F_c(u) = \frac{1}{H \cdot W} \sum_{i=1}^{H} \sum_{j=1}^{W} X(i,j,u),
\label{eq:channel_avg}
\end{equation}
where \( u \) indexes the channels. This descriptor is passed through a bottleneck layer with a reduction ratio \( r \) (set to 16), consisting of a dense layer with ReLU activation:
\begin{equation}
F_c' = \text{ReLU}(W_1 F_c),
\label{eq:channel_bottleneck}
\end{equation}
where \( W_1 \in \mathbb{R}^{C/r \times C} \). The output is then expanded and sigmoid-activated to produce channel weights \( M_c \in \mathbb{R}^{1 \times 1 \times C} \):
\begin{equation}
M_c = \sigma(W_2 F_c'),
\label{eq:channel_weight}
\end{equation}
where \( W_2 \in \mathbb{R}^{C \times C/r} \) and \( \sigma \) is the sigmoid function. The channel-refined feature \( X_c \) is obtained by element-wise multiplication:
\begin{equation}
X_c = X \otimes M_c.
\label{eq:channel_refine}
\end{equation}
\\
For spatial attention, average and max pooling are applied across channels to generate \( F_{avg} \in \mathbb{R}^{H \times W \times 1} \) and \( F_{max} \in \mathbb{R}^{H \times W \times 1} \):
\begin{equation}
F_{avg}(i,j) = \frac{1}{C} \sum_{u=1}^{C} X(i,j,u),
\label{eq:spatial_avg}
\end{equation}
\begin{equation}
F_{max}(i,j) = \max_{u=1}^{C} X(i,j,u).
\label{eq:spatial_max}
\end{equation}
These are concatenated along the channel dimension to form \( F_s \in \mathbb{R}^{H \times W \times 2} \), which is convolved with a 7$\times$7 kernel to produce a spatial attention map \( M_s \in \mathbb{R}^{H \times W \times 1} \):
\begin{equation}
M_s = \sigma(\text{Conv}_{7 \times 7}(F_s)).
\label{eq:spatial_map}
\end{equation}
The spatially refined feature \( X_s \) is computed as:
\begin{equation}
X_s = X \otimes M_s.
\label{eq:spatial_refine}
\end{equation}
The co-attention feature \( X_{co} \) is the element-wise product of \( X_c \) and \( X_s \):
\begin{equation}
X_{co} = X_c \otimes X_s.
\label{eq:co_attention}
\end{equation}
\\
\subsubsection*{Frequency Attention}
The frequency attention branch transforms \( X \) into the frequency domain using a 2D Fast Fourier Transform (FFT). The magnitude of the FFT, \( F_f \in \mathbb{R}^{H \times W \times C} \), is computed as:
\begin{equation}
F_f = |\text{FFT2D}(X)|,
\label{eq:fft}
\end{equation}
where \( |\cdot| \) denotes the absolute value. This magnitude is processed through two convolutional layers: a 1$\times$1 convolution to reduce channels to \( C/r \), followed by a 3$\times$3 convolution to restore \( C \) channels, both with ReLU activation:
\begin{equation}
F_f' = \text{ReLU}(\text{Conv}_{1 \times 1}(F_f)),
\label{eq:fft_reduce}
\end{equation}
\begin{equation}
F_f'' = \text{ReLU}(\text{Conv}_{3 \times 3}(F_f')),
\label{eq:fft_expand}
\end{equation}
A sigmoid activation generates the frequency attention map \( M_f \in \mathbb{R}^{H \times W \times C} \):
\begin{equation}
M_f = \sigma(F_f'').
\label{eq:freq_map}
\end{equation}
The frequency-refined feature \( X_f \) is:
\begin{equation}
X_f = X \otimes M_f.
\label{eq:freq_refine}
\end{equation}
\\
\subsubsection*{Dynamic Gating and Fusion}
The global average pooled feature \( F_g \in \mathbb{R}^{1 \times 1 \times C} \) from the co-attention module is used to compute a gating signal \( G \in \mathbb{R}^{1 \times 1 \times 1} \):
\begin{equation}
G' = \text{ReLU}(\text{Dense}_{32}(F_g)),
\label{eq:gate_dense1}
\end{equation}
\begin{equation}
G = \sigma(\text{Dense}_{1}(G')).
\label{eq:gate_dense2}
\end{equation}
Here, \( G \) is reshaped to broadcast across all spatial dimensions. The fused output \( X_{fuse} \) combines \( X_{co} \) and \( X_f \) weighted by \( G \) and \( 1 - G \):
\begin{equation}
X_{fuse} = (X_{co} \otimes G) + (X_f \otimes (1 - G)).
\label{eq:fusion}
\end{equation}
A residual connection adds the input \( X \) to \( X_{fuse} \):
\begin{equation}
X_{out} = X + X_{fuse}.
\label{eq:residual}
\end{equation}
\\
%%%%%%%%%%%%%%%%%%%%%%%%%%%%%%%%

\subsection{Proposed DB-FGA-Net Model}

The FGA block is integrated into the VGG16 and Xception backbones, pretrained on ImageNet. Let \( F_{vgg} \) and \( F_{xcep} \) denote the feature maps from VGG16 and Xception after FGA enhancement. These are concatenated:
\begin{equation}
F_{concat} = \text{Concat}(F_{vgg}, F_{xcep}),
\label{eq:concat}
\end{equation}
followed by a 1$\times$1 convolution to produce \( F_{fuse} \in \mathbb{R}^{H \times W \times 1024} \):
\begin{equation}
F_{fuse} = \text{ReLU}(\text{Conv}_{1 \times 1}(F_{concat})).
\label{eq:fuse_conv}
\end{equation}
Global average pooling reduces \( F_{fuse} \) to \( F_{pool} \in \mathbb{R}^{1 \times 1 \times 1024} \), followed by dropout with rate 0.3:
\begin{equation}
F_{drop} = \text{Dropout}_{0.3}(F_{pool}).
\label{eq:dropout}
\end{equation}
The final classification is performed with a dense layer and softmax:
\begin{equation}
P = \text{Softmax}(\text{Dense}_{4}(F_{drop})),
\label{eq:softmax}
\end{equation}
where \( P \in \mathbb{R}^{1 \times 4} \) represents the class probabilities. The model is trained using Adam optimizer with learning rate \( 1 \times 10^{-4} \) and categorical cross-entropy loss over 20 epochs, with early stopping.

\begin{table}[h]
\centering
% \small
\caption{Parameter breakdown of DB-FGA-Net}
\label{tab:db-fga-net-params}
\renewcommand{\arraystretch}{1.5} % row height
\setlength{\tabcolsep}{6pt} % column padding
\label{tab:training_params}
\begin{tabular}{|l|l|l|}
\hline
\textbf{Component} & \textbf{Output Shape} & \textbf{Parameters} \\
\hline
VGG16 + FGA block          & (None, 8, 8, 512)   & 17,387,012 \\
Xception + FGA block       & (None, 8, 8, 2048)  & 63,400,780 \\
Feature Concatenation      & (None, 8, 8, 2560)  & 0          \\
1$\times$1 Fusion Conv (ReLU) & (None, 8, 8, 1024) & 2,622,464  \\
Global Average Pooling     & (None, 1024)        & 0          \\
Dropout (0.3)              & (None, 1024)        & 0          \\
Dense + Softmax (4 classes)& (None, 4)           & 4,100      \\
\hline
\textbf{Total parameters}  & --                  & \textbf{83,414,356} \\
Trainable                  & --                  & 83,359,828 \\
Non-trainable              & --                  & 54,528 \\
\hline
\end{tabular}
\end{table}

The DB-FGA-Net architecture contains approximately 83.4 million parameters, with the overwhelming majority ($\sim$80.8M) coming from the pretrained VGG16 and Xception backbones. The proposed FGA blocks, fusion layer and classification head add only a modest number of parameters ($\sim$2.6M), making the model computationally efficient relative to its representational capacity. This design balances high feature expressiveness due to dual complementary backbones and frequency guided attention with reasonable training and inference costs.

\subsection{Training Setup}

\begin{table}[h]
\centering
\caption{Training Hyperparameters and Setup Details}
% Adjust table spacing
\renewcommand{\arraystretch}{1.5} % row height
\setlength{\tabcolsep}{6pt} % column padding
\label{tab:training_params}
\begin{tabular}{|l|l|}
% \toprule
\hline
\textbf{Parameter} & \textbf{Value}\\ \hline
% \midrule
Optimizer & Adam \\ \hline
Learning Rate & $10^{-4}$ \\ \hline
Batch Size & 32 \\ \hline
Epochs & 30 or 20 \\ \hline
Loss Function & Categorical Cross-Entropy \\ \hline
Hardware & NVIDIA P100 GPU (Kaggle) \\ \hline
% \bottomrule
\end{tabular}
\end{table}

The training pipeline uses the 7K-DS dataset, stratified into 80\% training and 20\% validation subsets to preserve class balance. Images are resized to $256 \times 256$ pixels using bicubic interpolation and normalized to the $[0, 1]$ range:
$$I_{\text{norm}}(x, y, c) = \frac{I_{\text{resized}}(x, y, c)}{255},$$
where $c \in {R, G, B}$ denotes color channels. Training employs the Adam optimizer with a learning rate of $10^{-4}$, a batch size of 32 and 30 epochs for 4 and 3-class, 20 epochs for 2-class dataset. The loss function is categorical cross-entropy, suitable for multi-class tasks with one-hot encoded labels $y_i$:
$$\mathcal{L} = -\sum_{i=1}^{N} \sum_{k=1}^{C} y_{i,k} \log(p_{i,k}),$$
where $N$ is the batch size, $C$ is the number of classes, and $p_{i,k}$ is the predicted probability. Five-fold cross-validation is executed, with each fold trained independently on a Kaggle environment using an NVIDIA P100 GPU, ensuring statistical robustness.
The key training parameters are summarized in Table~\ref{tab:training_params}.

\subsection{Evaluation Metrics}

\begin{table*}[h]
\centering
\caption{Quantitative results on 7K-DS (4-Class classification).}
% Adjust table spacing
\renewcommand{\arraystretch}{1.5} % row height
\setlength{\tabcolsep}{6pt} % column padding
\label{tab:4class_results}
\begin{tabular}{|l|c|c|c|c|}
\hline
\textbf{Model}          & \textbf{Accuracy} & \textbf{Precision} & \textbf{Recall} & \textbf{F1-Score} \\ \hline
VGG16                   & 96.19\%           & 95.98\%            & 95.20\%         & 96.08\%           \\ 
VGG19                   & 95.50\%           & 96.04\%            & 95.50\%         & 95.57\%           \\ 
MobileNetV2             & 86.65\%           & 90.17\%            & 86.65\%         & 87.10\%           \\ 
Xception                & 97.94\%           & 98.00\%            & 97.94\%         & 97.94\%           \\ \hline
VGG16 + FGA            & 98.40\%           & 98.40\%            & 98.40\%         & 98.39\%           \\ 
VGG19 + FGA            & 97.86\%           & 97.92\%            & 97.86\%         & 97.87\%           \\ 
MobileNetV2 + FGA      & 97.48\%           & 97.52\%            & 97.48\%         & 97.48\%           \\ 
Xception + FGA         & 98.40\%           & 98.43\%            & 98.40\%         & 98.40\%           \\ \hline
\textbf{DB-FGA-Net (Proposed)} & \textbf{99.24\%} & \textbf{99.24\%} & \textbf{99.24\%} & \textbf{99.24\%} \\ \hline
\end{tabular}
\end{table*}

% \begin{figure*}[h]
%     \centering
%     \begin{subfigure}[b]{0.95\textwidth}
%         \includegraphics[width=\textwidth]{images/fusion_4c.png}
%         \caption{4-Class Training \& Validation Accuracy/Loss Curves}
%     \end{subfigure}\\
%     % \hfill
%     \begin{subfigure}[b]{0.95\textwidth}
%         \includegraphics[width=\textwidth]{images/fusion_3c.png}
%         \caption{3-Class Training \& Validation Accuracy/Loss Curves}
%     \end{subfigure}\\
%     % \hfill
%     \begin{subfigure}[b]{0.95\textwidth}
%         \includegraphics[width=\textwidth]{images/fusion_2c.png}
%         \caption{2-Class Training \& Validation Accuracy/Loss Curves}
%     \end{subfigure}
%     \caption{Training and validation accuracy/loss curves for the dual-backbone feature fusion model on 7K-DS across 4-class, 3-class, and 2-class settings.}
%     \label{fig:dual_backbone_training}
% \end{figure*}

\begin{figure*}[h]
    \centering
    \begin{subfigure}[b]{0.45\textwidth}
        \centering
        \includegraphics[width=\textwidth]{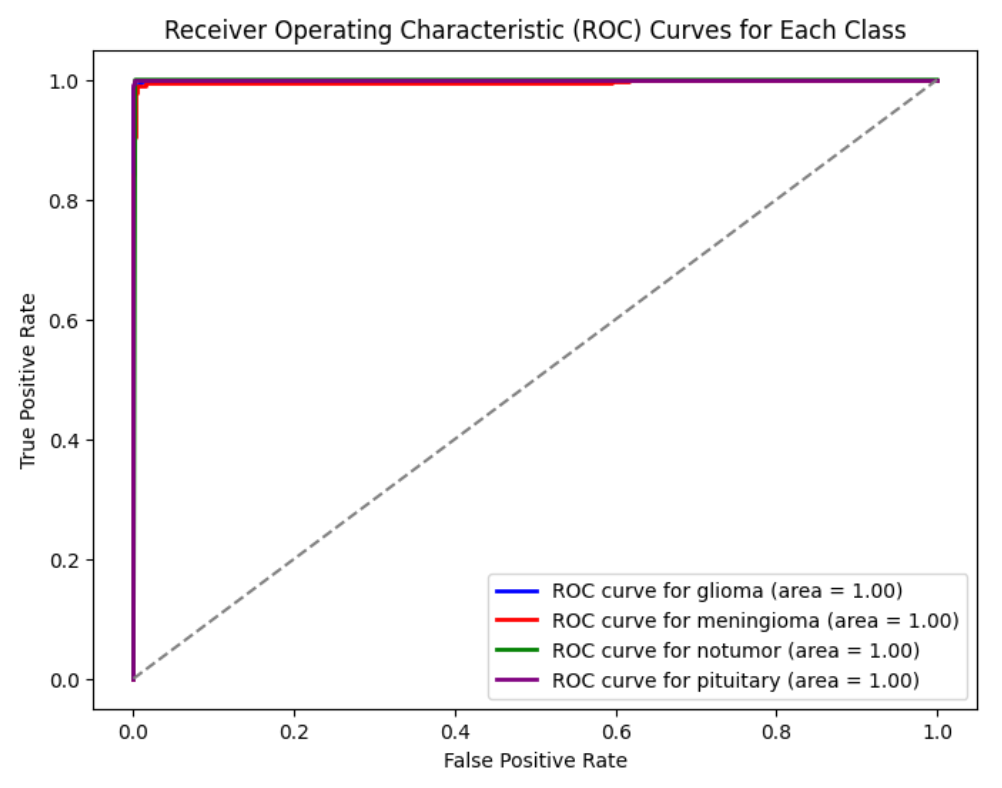}
        \caption{ROC Curve}
    \end{subfigure}
    \hfill
    \begin{subfigure}[b]{0.45\textwidth}
        \centering
        \includegraphics[width=\textwidth]{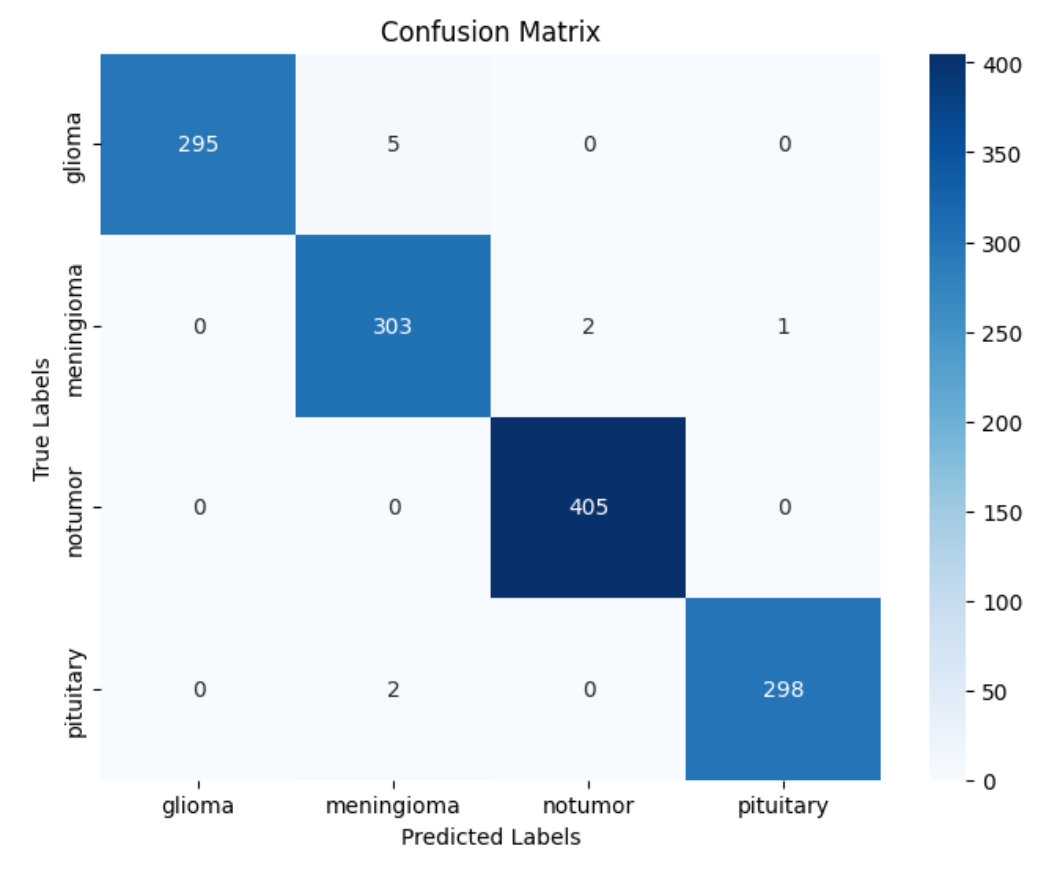}
        \caption{Confusion Matrix}
    \end{subfigure}
    \caption{4-Class ROC and Confusion Matrix for the proposed model on 7K-DS.}
    \label{fig:4c_roc_confusion}
\end{figure*}

\begin{figure*}[h]
    \centering
    \begin{subfigure}[b]{0.45\textwidth}
        \centering
        \includegraphics[width=\textwidth]{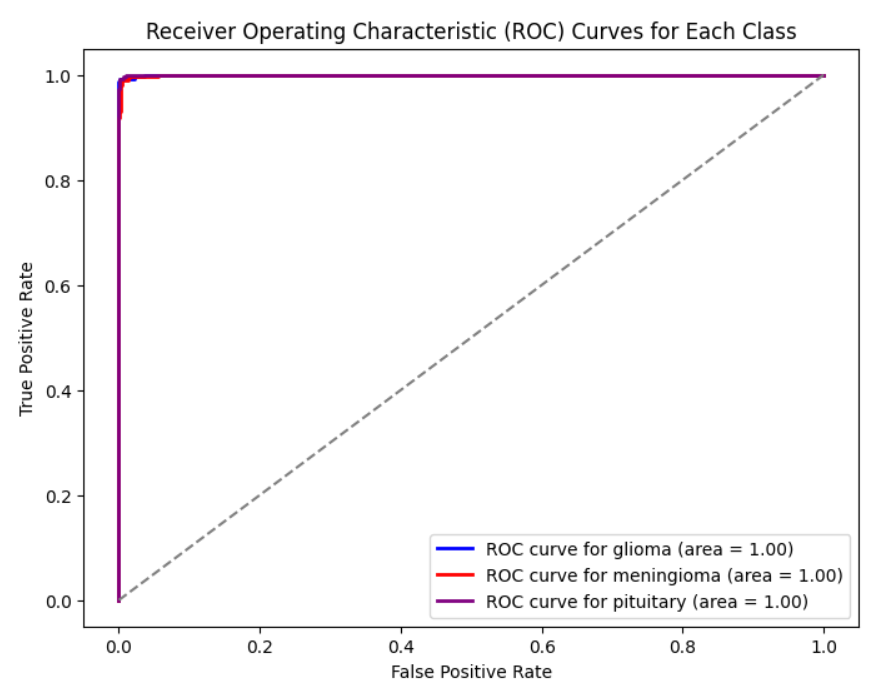}
        \caption{ROC Curve}
    \end{subfigure}
    \hfill
    \begin{subfigure}[b]{0.45\textwidth}
        \centering
        \includegraphics[width=\textwidth]{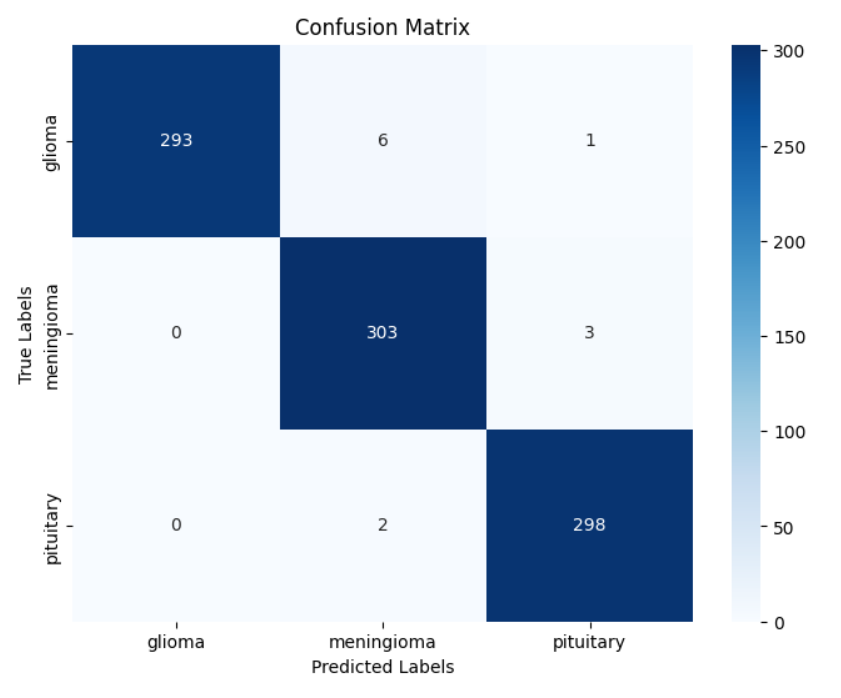}
        \caption{Confusion Matrix}
    \end{subfigure}
    \caption{3-Class ROC and Confusion Matrix for the proposed on 7K-DS.}
    \label{fig:3c_roc_confusion}
\end{figure*}

\begin{figure*}[h]
    \centering
    \begin{subfigure}[b]{0.45\textwidth}
        \centering
        \includegraphics[width=\textwidth]{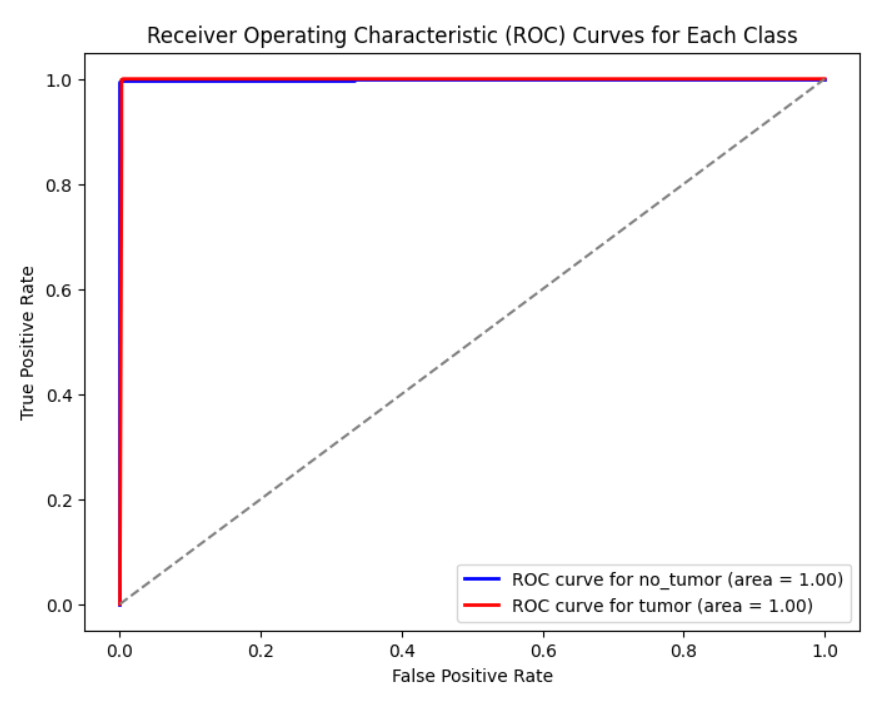}
        \caption{ROC Curve}
    \end{subfigure}
    \hfill
    \begin{subfigure}[b]{0.45\textwidth}
        \centering
        \includegraphics[width=\textwidth]{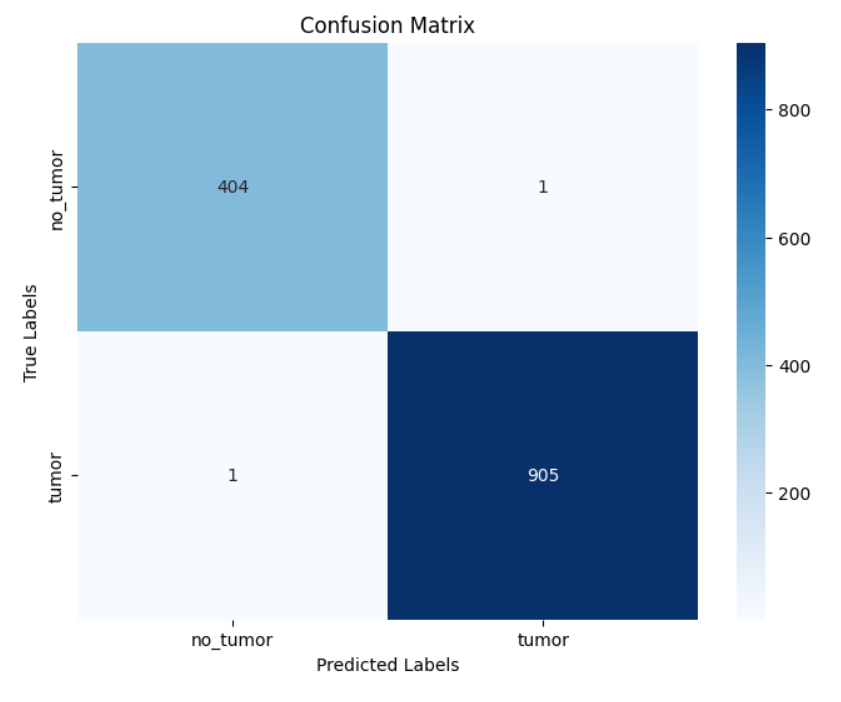}
        \caption{Confusion Matrix}
    \end{subfigure}
    \caption{2-Class ROC and Confusion Matrix for the proposed model on 7K-DS.}
    \label{fig:2c_roc_confusion}
\end{figure*}

\begin{figure*}[h]
    \centering
    \begin{subfigure}[b]{0.48\textwidth}
        \centering
        \includegraphics[width=\textwidth]{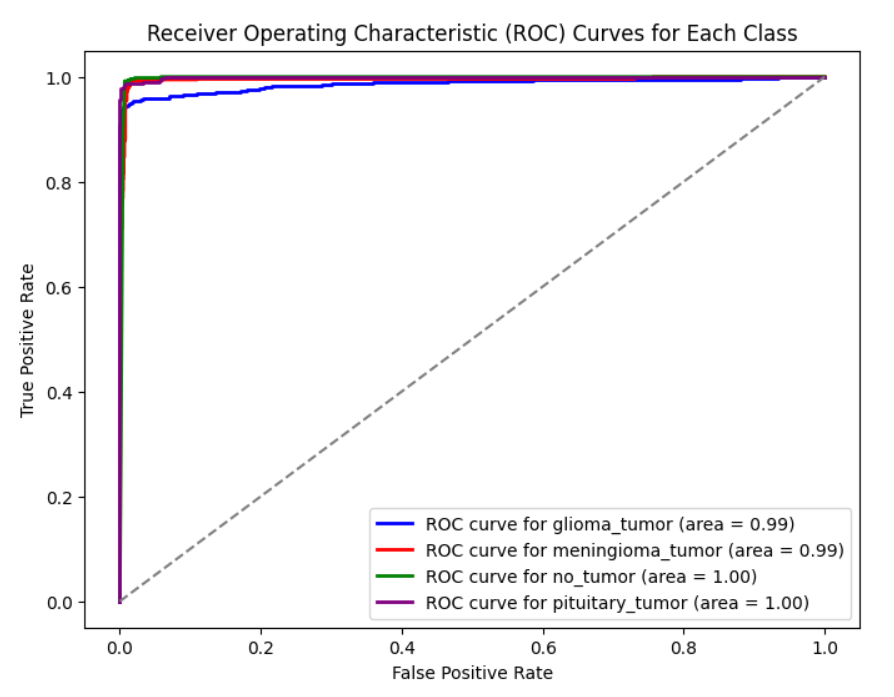}
        \caption{4-Class ROC Curve}
    \end{subfigure}
    \hfill
    \begin{subfigure}[b]{0.48\textwidth}
        \centering
        \includegraphics[width=\textwidth]{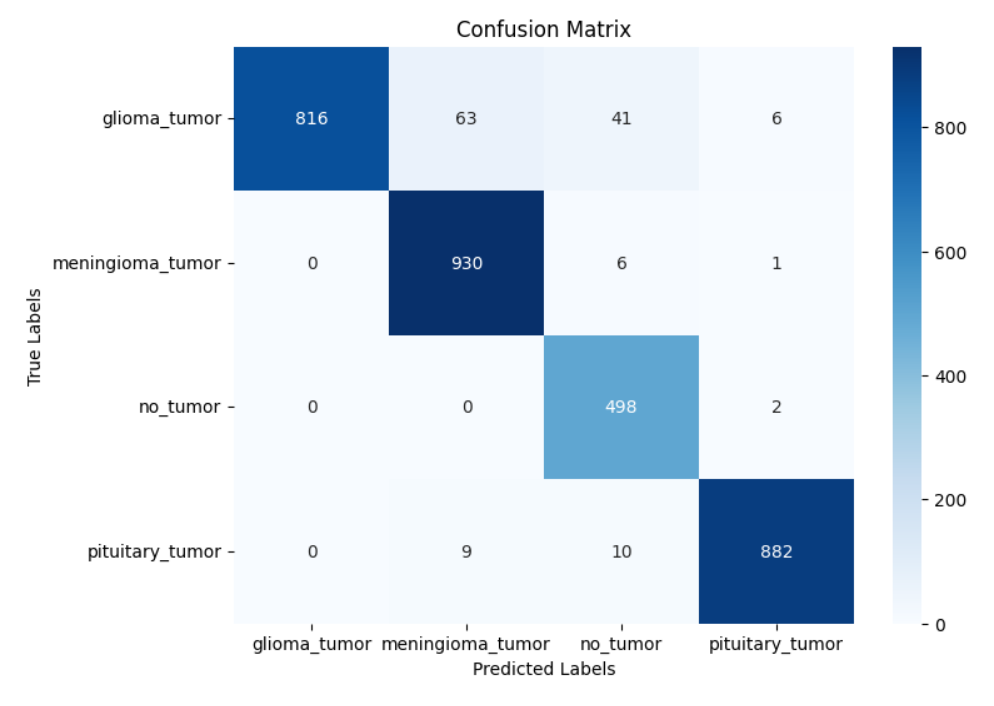}
        \caption{4-Class Confusion Matrix}
    \end{subfigure}
    \caption{ROC curve and confusion matrix for the proposed model validated on 3K-DS, illustrating discriminative power and per-class performance using weights trained on 7K-DS.}
    \label{fig:dual_backbone_3kds_confusion}
\end{figure*}

\begin{table*}[h]
\centering
\renewcommand{\arraystretch}{1.5}
\setlength{\tabcolsep}{6pt}
\caption{Performance Comparison with CBAM and FGA on 7K-DS (4-Class)}
\label{tab:cbam_comparison}
\begin{tabular}{|l|c|c|c|c|}
\hline
\textbf{Models} & \textbf{Accuracy} & \textbf{Precision} & \textbf{Recall} & \textbf{F1-Score} \\
\hline
\textbf{VGG16 + FGA} & \textbf{98.40\%} & \textbf{98.40\%} & \textbf{98.40\%} & \textbf{98.39\%} \\
VGG16 + CBAM & 97.48\% & 97.5\% & 97.48\% & 97.48\% \\
\hline
\textbf{VGG19 + FGA} & \textbf{97.86\%} & \textbf{97.92\%} & \textbf{97.86\%} & \textbf{97.87\%} \\
VGG19 + CBAM & 97.71\% & 97.73\% & 97.71\% & 97.71\% \\
\hline
\textbf{MobileNetV2 + FGA} & \textbf{97.48\%} & \textbf{97.52\%} & \textbf{97.48\%} & \textbf{97.48\%} \\
MobileNetV2 + CBAM & 85.43\% & 89.26\% & 85.43\% & 85.67\% \\
\hline
Xception + FGA & 98.40\% & 98.43\% & 98.40\% & 98.40\% \\
\textbf{Xception + CBAM} & \textbf{98.55\%} & \textbf{98.61\%} & \textbf{98.55\%} & \textbf{98.55\%} \\
\hline
\end{tabular}
\end{table*}

To comprehensively assess the performance of the brain tumor classification models, a suite of standard classification metrics is employed, complemented by visual diagnostic tools. The primary metrics include accuracy, precision, recall, and F1-score, defined as follows:
\begin{align}
\text{Accuracy} &= \frac{TP + TN}{TP + TN + FP + FN}, \\
\text{Precision} &= \frac{TP}{TP + FP}, \\
\text{Recall} &= \frac{TP}{TP + FN}, \\
\text{F1-Score} &= 2 \cdot \frac{\text{Precision} \cdot \text{Recall}}{\text{Precision} + \text{Recall}},
\end{align}
where $TP$, $TN$, $FP$, and $FN$ represent true positives, true negatives, false positives, and false negatives, respectively. For multi-class scenarios (4-class, 3-class and 2-class settings), these metrics are computed per class and macro-averaged to provide an overall performance assessment across the glioma, meningioma, no tumor, and pituitary categories.
\\
Additionally, confusion matrices are generated to provide a detailed breakdown of model predictions. It offers insights into class-specific performance and misclassification patterns. Each matrix is a $C \times C$ grid, where $C$ is the number of classes, with diagonal elements representing correct predictions (e.g., $TP$ for each class) and off-diagonal elements indicating misclassifications (e.g., $FP$ or $FN$). This visualization, evaluated on the 7K-DS validation set and 3K-DS test set, aids in identifying potential biases or errors, such as confusion between similar tumor types.
\\
Further, receiver operating characteristic (ROC) curves are generated using the one-vs-rest (OvR) strategy for multi-class evaluation, with the area under the curve (AUC) quantifying the model's discriminative ability:
\begin{equation}
\text{AUC} = \int_0^1 \text{TPR}(t) \, d\text{FPR}(t),
\end{equation}
where TPR (true positive rate) is $\frac{TP}{TP + FN}$ and FPR (false positive rate) is $\frac{FP}{FP + TN}$ at threshold $t$. The AUC provides a threshold-independent measure of performance, assessed across all datasets to ensure generalization. These metrics collectively enable a robust evaluation of the model's effectiveness in classifying brain tumors across diverse settings.
\\
\section{Result Analysis}

The experimental results highlight the performance of the proposed FGA-enhanced dual backbone framework (DB-FGA-Net) for brain tumor classification. The models were evaluated across 4-class, 3-class and 2-class settings on the 7K-DS dataset, with additional cross-dataset generalization testing on the 3K-DS dataset. The framework achieves high performance metrics without data augmentation, demonstrating its robustness and improved performance compared to baseline and selected existing approaches. Validation is supported by 5-fold cross-validation, confusion matrices, ROC curves, AUC analyses, and most importantly Grad-CAM visualizations, which enhance interpretability.  

\subsection{Performance on 7K-DS Dataset}

\begin{table*}[htbp]
\centering
\caption{5-fold cross-validation results of the proposed model.}
\renewcommand{\arraystretch}{1.5}
\setlength{\tabcolsep}{6pt}
\label{tab:5fold_results}
\begin{tabular}{|c|c|c|c|c|}
\hline
\textbf{Fold} & \textbf{Accuracy (\%)} & \textbf{Macro Precision (\%)} & \textbf{Macro Recall (\%)} & \textbf{Macro F1-score (\%)} \\
\hline
Fold 1 & 98.47 & 98.42 & 98.41 & 98.41 \\
Fold 2 & 98.86 & 98.82 & 98.76 & 98.79 \\
Fold 3 & 98.09 & 97.99 & 98.01 & 98.10 \\
Fold 4 & 98.74 & 98.39 & 98.43 & 98.41 \\
Fold 5 & 99.24 & 99.20 & 99.24 & 99.24 \\
\hline
\textbf{Mean} & \textbf{98.68} & \textbf{98.56} & \textbf{98.57} & \textbf{98.59} \\
\hline
\end{tabular}
\end{table*}

\subsubsection{Quantitative Evaluation}

\begin{table}[h]
\centering
\caption{Quantitative Results on 7K-DS (3-Class)}
\renewcommand{\arraystretch}{1.5}
\setlength{\tabcolsep}{6pt}
\label{tab:3class_results}
\begin{tabular}{|l|c|c|c|c|}
\hline
\textbf{Model} & \textbf{Accuracy} & \textbf{Precision} & \textbf{Recall} & \textbf{F1-Score} \\ \hline
Proposed & 98.68\% & 98.69\% & 98.68\% & 98.68\% \\ \hline
\end{tabular}
\end{table}

\begin{table}[h]
\centering
\caption{Quantitative Results on 7K-DS (2-Class)}
\renewcommand{\arraystretch}{1.5}
\setlength{\tabcolsep}{6pt}
\label{tab:2class_results}
\begin{tabular}{|l|c|c|c|c|}
\hline
\textbf{Model} & \textbf{Accuracy} & \textbf{Precision} & \textbf{Recall} & \textbf{F1-Score} \\ \hline
Proposed & 99.85\% & 99.85\% & 99.85\% & 99.85\% \\ \hline
\end{tabular}
\end{table}

The 7K-DS dataset represents the primary benchmark used for evaluation. It contains MRI images from four categories: glioma, meningioma, no tumor, and pituitary. This diversity of classes makes it an excellent testbed for evaluating how well a deep learning model can separate subtle tumor variations, as well as distinguish tumor from non-tumor conditions.  
\\
For the 4-class experiment, the proposed DB-FGA-Net outperformed the evaluated baseline and single-backbone variants, achieving 99.24\% accuracy along with equally high precision, recall, and F1-score. This level of consistency across multiple evaluation metrics is a strong indicator that the network is not biased toward any particular tumor type.   
\\
When the problem is simplified to 3-class and 2-class tasks, accuracies remain consistently high, with 98.68\% and 99.85\% respectively. The gradual rise in accuracy as the classification task becomes less complex suggests that the model generalizes its learned features well, even when classes are merged. This reflects what is often seen in medical imaging studies, where binary decisions (tumor vs. no tumor) are easier than fine-grained multi-class separation. However, what sets our method apart is that the drop from binary to multi-class performance is modest, implying strong representational power even for difficult discrimination between tumor subtypes.  
\\
These results are particularly notable because they were achieved without data augmentation. In the literature, augmentation (rotation, flipping, noise injection, etc.) is frequently used to artificially increase diversity and avoid overfitting. Achieving strong results without such augmentation demonstrates the inherent robustness of our feature extraction strategy and its ability to learn meaningful spectral-spatial representations directly from available data.  
\\
\subsubsection{5-Fold Cross-Validation}
To further evaluate the reliability of the proposed framework, a 5-fold cross-validation strategy was implemented. Cross-validation is essential for models reliability because dataset splits can heavily influence outcomes when the sample size is relatively limited. By averaging over multiple folds, the performance variance due to random splits is minimized, providing a more trustworthy estimate of real-world performance.
\\
The accuracies across folds range from 98.09\% to 99.24\%, with a standard deviation below 0.5\%. This low variability suggests that the model is not overly sensitive to the specific composition of training and validation splits. In practical terms, this implies that clinicians could expect the model to maintain high accuracy even if applied to new hospital datasets with slightly different patient distributions. This property is highly desirable in medical AI, where overfitting to a particular dataset could have serious implications for generalizability.  
\\
\subsubsection{ROC and AUC Analysis}
Receiver Operating Characteristic (ROC) curves and Area Under the Curve (AUC) metrics provide a more nuanced view of classifier performance beyond simple accuracy. An AUC of 1.00, achieved in all classes for the 4-class and 3-class settings, indicates perfect discrimination between positive and negative examples for each class. Even in the binary setting, the ROC curve was almost ideal, reflecting the network’s ability to separate tumor from non-tumor cases with virtually no false positives or false negatives.  
\\
This result is particularly encouraging in medical applications where false negatives (missed tumors) could lead to life-threatening consequences, and false positives could trigger unnecessary biopsies or treatments. By demonstrating a ROC-AUC of 1.00 across categories, the model positions itself as not only accurate but also clinically safe.  
\\
\subsubsection{Confusion Matrices}
While overall accuracy is important, confusion matrices reveal per-class strengths and weaknesses. In the 4-class task, small misclassifications were observed between glioma and meningioma. This is understandable since these tumors share overlapping textural characteristics in MRI scans, and even radiologists sometimes face challenges distinguishing them. However, classes such as “no tumor” and “pituitary” were almost perfectly identified.  
\\
In the binary setting, the confusion matrix revealed nearly flawless separation between tumor and non-tumor cases, with negligible misclassification rates. This robustness across different granularities strengthens confidence in the model’s clinical usability, showing that its predictions align with medically meaningful distinctions.

\subsection{Cross-Dataset Validation (3K-DS)}
To assess the real-world applicability and robustness of the dual-backbone feature fusion model, we evaluated DB-FGA-Net on the independent 3K-DS dataset. Unlike the 7K-DS training set, this dataset contains images that may differ in acquisition protocols, scanner settings, and patient demographics, thereby simulating a realistic domain shift. Testing under such conditions is crucial in medical AI because a model that performs well only on a single curated dataset may fail in clinical practice where variations are inevitable.  

\begin{table}[h]
\centering
\caption{Quantitative Results on 3K-DS}
\renewcommand{\arraystretch}{1.5}
\setlength{\tabcolsep}{6pt}
\label{tab:3kds_results}
\begin{tabular}{|l|c|c|c|c|}
\hline
\textbf{Model} & \textbf{Accuracy} & \textbf{Precision} & \textbf{Recall} & \textbf{F1-Score} \\ \hline
Proposed & 95.77\% & 96.01\% & 95.77\% & 95.75\% \\ \hline
\end{tabular}
\end{table}

The proposed framework achieved a 4-class accuracy of 95.77\%, with precision at 96.01\%, recall at 95.77\%, and F1-score at 95.75\% (Table~\ref{tab:3kds_results}). While this marks a modest performance drop of about 3.5\% compared to results on 7K-DS, it is important to note that such degradation is expected when transitioning across domains. What is significant is that the model still maintains above 95\% accuracy without retraining or fine-tuning, demonstrating its strong generalization capability.

\begin{table}[htbp]
\centering
\caption{Optimizer and Hyperparameter Sensitivity Analysis of the Proposed Model on 7K-DS (4-Class).}
\label{tab:optimizer_comparison}
\renewcommand{\arraystretch}{1.5}
\setlength{\tabcolsep}{6pt}
\begin{tabular}{|c|c|c|c|c|c|c|}
\hline
\textbf{Optimizer} & \textbf{Batch Size} & \textbf{LR} & \textbf{Acc (\%)} & \textbf{Macro Pre (\%)} & \textbf{Macro Recall (\%)} & \textbf{Macro F1-Score (\%)} \\ \hline
Adamax    & 16  & $1\times10^{-5}$ & 98.25 & 98.19 & 98.10 & 98.14 \\ \hline
Adamax    & 16  & $1\times10^{-4}$ & 98.09 & 98.03 & 97.98 & 98.00 \\ \hline
Adamax    & 32  & $1\times10^{-5}$ & 97.71 & 97.67 & 97.53 & 97.59 \\ \hline
Adamax    & 32  & $1\times10^{-4}$ & 97.41 & 97.31 & 97.25 & 97.27 \\ \hline
SGD       & 16  & $1\times10^{-5}$ & 52.63 & 53.20 & 49.51 & 46.50 \\ \hline
SGD       & 16  & $1\times10^{-4}$ & 90.39 & 90.09 & 90.04 & 90.00 \\ \hline
Adam      & 16  & $1\times10^{-5}$ & 97.56 & 97.58 & 97.35 & 97.43 \\ \hline
Adam      & 16  & $1\times10^{-4}$ & 98.86 & 98.80 & 98.85 & 98.82 \\ \hline
Adam      & 32  & $1\times10^{-5}$ & 96.95 & 96.92 & 96.71 & 96.77 \\ \hline
Adam      & 32  & $1\times10^{-4}$ & \textbf{99.24} & \textbf{99.23} & \textbf{99.17} & \textbf{99.20} \\ \hline
\end{tabular}
\end{table}

\section{Discussion}

\subsection{Ablation Study}
To evaluate the contribution of the Frequency-Gated Attention (FGA) and dual-backbone design, we conducted ablation experiments summarized in Table~\ref{tab:4class_results} and Table~\ref{tab:cbam_comparison}.
\\
As shown in Table~\ref{tab:4class_results}, introducing FGA to backbone models yielded consistent improvements. For instance, VGG16 improved from 96.19\% to 98.40\%, and MobileNetV2 from 86.65\% to 97.48\%. The proposed DB-FGA-Net achieved the highest performance with 99.24\% accuracy, surpassing all single-backbone baselines. Table~\ref{tab:cbam_comparison} further compares FGA with the widely used CBAM module. FGA demonstrated superior performance across most backbones, with improvements up to 13.81\%. While CBAM slightly outperformed FGA on Xception (98.55\% vs. 98.40\%), the proposed DB-FGA-Net with FGA achieved overall more stable and superior results.  
\\
These ablations confirm two important points: (i) frequency gating is crucial for capturing both high-frequency tumor boundaries and low-frequency contextual information, and (ii) the dual-backbone design enhances robustness by combining complementary feature representations.

\begin{table}[H]
\centering
\caption{Comprehensive Comparison of Brain Tumor Classification Models}
\label{tab:model_comparison}

% Adjust table spacing
\renewcommand{\arraystretch}{1.5} % row height
\setlength{\tabcolsep}{6pt} % column padding

\resizebox{\textwidth}{!}{%
\begin{tabular}{|p{1.2cm}|p{1.2cm}|p{2.4cm}|p{3.3cm}|p{1.5cm}|p{1.5cm}|p{1.5cm}|}
\hline
\textbf{Authors} & \textbf{Dataset}  & \textbf{Model} & \textbf{Accuracy} & \textbf{Precision} & \textbf{Recall} & \textbf{F1-Score} \\ \hline

\cite{PBviT} & Figshare (3 class)  & Patch-Based Vision Transformer & 95.80\% & 95.30\% & 93.20\% & 92\% \\ \hline

\cite{GGLA} & 7K-DS, 

\vspace{1cm} Figshare (3 class)  & Dual-Branch Gated Attention Network & 96.80\%7(7K-DS, No-aug), 96.50\%(7K-DS, aug), 99.62\%(7K-DS, GAN-aug),   

\vspace{0.5cm}

96.42\%(Figshare, no aug),
98.58\%(Figshare, sim aug),
99.06\%(Figshare, GAN aug)

& 96.75\%, 99.48\%, 99.60\%,   

\vspace{0.5cm}

96.32\%,
98.59\%,
99.05\% &

96.76\%, 99.50\%, 99.62\%,   

\vspace{0.5cm}

95.57\%,
98.59\%,
99.05\%& 

96.73\%, 96.48\%, 99.62\%,   

\vspace{0.5cm}

96.02\%,
98.59\%,
99.06\% \\ \hline

\cite{FedAvg}  & 7K-DS  & CNN + FedAvg + FedProx & 97.19\% & High & High & 97.18\% \\ \hline

\cite{trans}  & 7K-DS & ViT + EfficientNetV2 (Ensemble) & 96\% & 96\% & 96\% & 96\% \\ \hline

\cite{hybrid3b}  & Combined 4 datasets  & Hybrid 3B Net + EfficientNetB2 & 97.80\% (4-class), 

98.72\% (3-class), 

99.50\% (2-class) & High & High & High \\ \hline

\cite{fuzzy}  & 7K-DS, 

3K-DS, 

13K-DS & Fuzzy Thresholding + DL & 

98.42\% (7K-DS),

97.22\% (3K-DS),

98.18\% (13K-DS)

& 

98.32\%,

98.16\%,

99.42\%

 & 
 98.14\%,

97.21\%,

98.26\%
 & 
 
 98.10\%,

98.11\%,

98.65\%
\\ \hline

\cite{deep} & 7K-DS, 3K-DS & EfficientNetB0 w/ Dual Reg. & 98\% & 95\% & 98.2\% & 95.4\% \\ \hline

\cite{hybrid-cnn}  & 7K-DS & ResNet50V2 + MobileNetV2 + DenseNet121 & 98.75\% & 98.76\% & 98.75\% & 98.75\% \\ \hline

\textbf{Proposed} & 7K-DS, 

\vspace{1cm}
3K-DS & DB-FGA-Net & 
99.24\% (7K-DS, 4-class), 98.68\% (7K-DS, 3-class), 99.85\% (7K-DS, 2-class),

\vspace{0.5cm}
95.77\% (3K-DS, 4-class) & 
99.24\%, 98.69\%, 99.85\%, 

\vspace{0.5cm}
96.01\% & 
99.24\%, 98.68\%, 99.85\%, 

\vspace{0.5cm}
95.77\% & 
99.24\%, 98.68\%, 99.85\%, 

\vspace{0.5cm}
95.75\% \\ \hline

\end{tabular}
}
\end{table}

\subsection{Optimizer and Hyperparameter Sensitivity Analysis}
To investigate the effect of training configurations on the performance of DB-FGA-Net, we experimented with three optimizers (Adam, Adamax, SGD), two batch sizes (16, 32), and two learning rates ($1 \times 10^{-5}$ and $1 \times 10^{-4}$). The results are summarized in Table~\ref{tab:optimizer_comparison}.
\\
The findings indicate that the choice of optimizer has the most significant impact. Adam consistently provided the most stable and accurate results, with its best setting (batch size of 32, learning rate of $1 \times 10^{-4}$) achieving the highest overall accuracy of 99.24\% along with balanced precision, recall, and F1-score values above 99\%. This demonstrates the optimizer's ability to converge efficiently and maintain robust feature learning. Adam also showed strong performance at smaller batch sizes and lower learning rates, though slightly below its best configuration.
\\
Adamax delivered competitive results (97.4-98.3\% accuracy), showing stable behavior across learning rates and batch sizes, but consistently underperforming compared to Adam. This suggests that while Adamax can be a viable alternative, its adaptive learning scheme is slightly less suited to the spectral-spatial complexity of MRI features.
\\
By contrast, SGD struggled significantly, especially at the lower learning rate ($1 \times 10^{-5}$), where performance collapsed to 52.6\%. Even at the higher learning rate, accuracy peaked at only 90.39\%, far below Adam and Adamax. This highlights SGD's sensitivity to parameter tuning and its limited effectiveness in capturing the fine-grained frequency and texture cues critical for this task.
\\
Overall, the experiments confirm that: (i) Adam is the optimal optimizer for DB-FGA-Net, particularly with a learning rate of $1 \times 10^{-4}$ and a moderate batch size (32), (ii) Adamax is stable but less effective, and (iii) SGD is not well suited to this application. The batch size showed only minor influence compared to the optimizer choice, reinforcing that optimization strategy is the primary driver of performance.

\subsection{Grad-CAM Visualization and Interpretability}
\begin{figure}[H]
    \centering
    \includegraphics[width=0.96\textwidth]{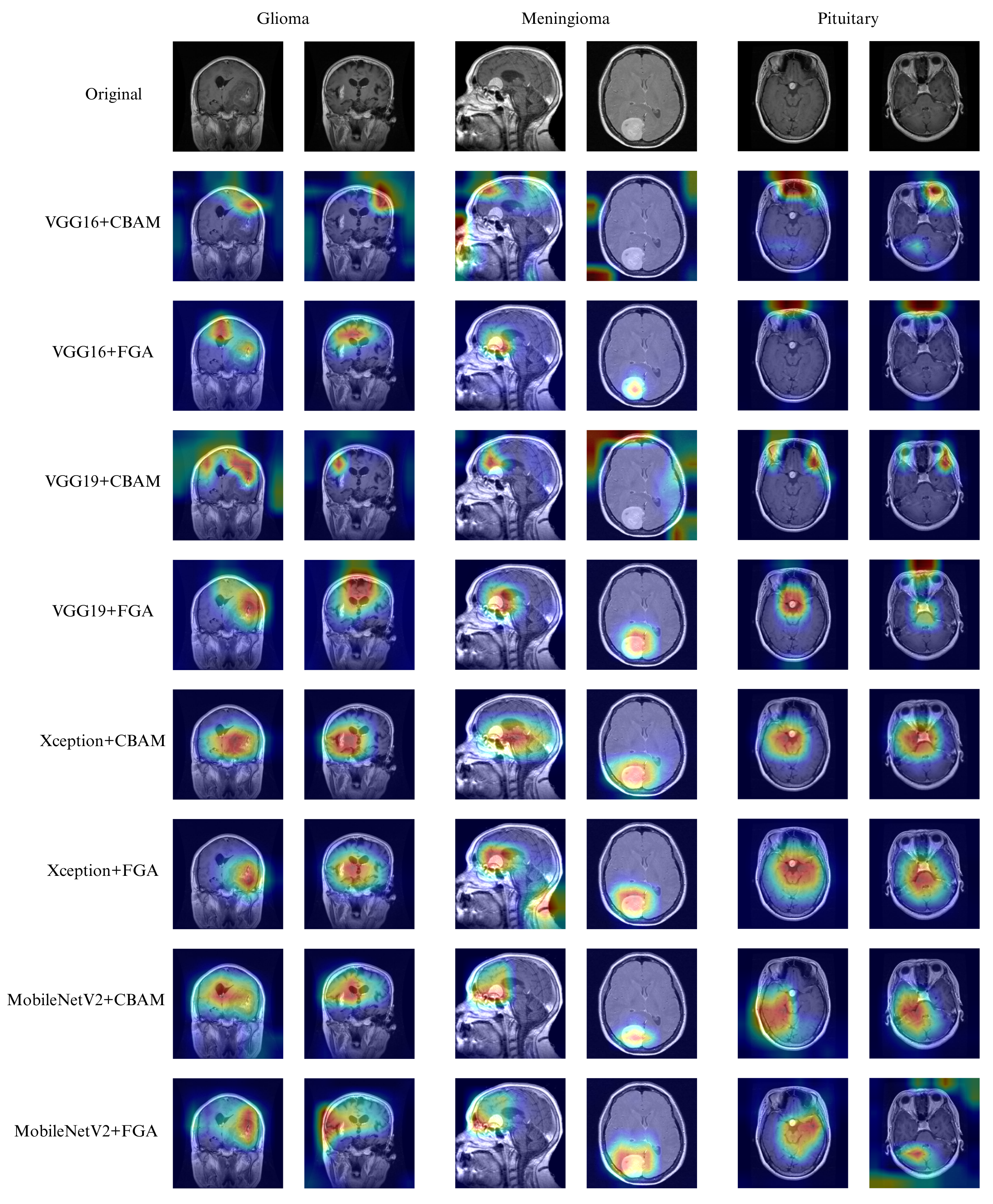}
    \caption{Grad-CAM comparison of the CBAM and FGA integrated models on 7K-DS (4-class) }
    \label{fig:6}
\end{figure}

\begin{figure}[H]
    \centering
    \includegraphics[width=0.96\textwidth]{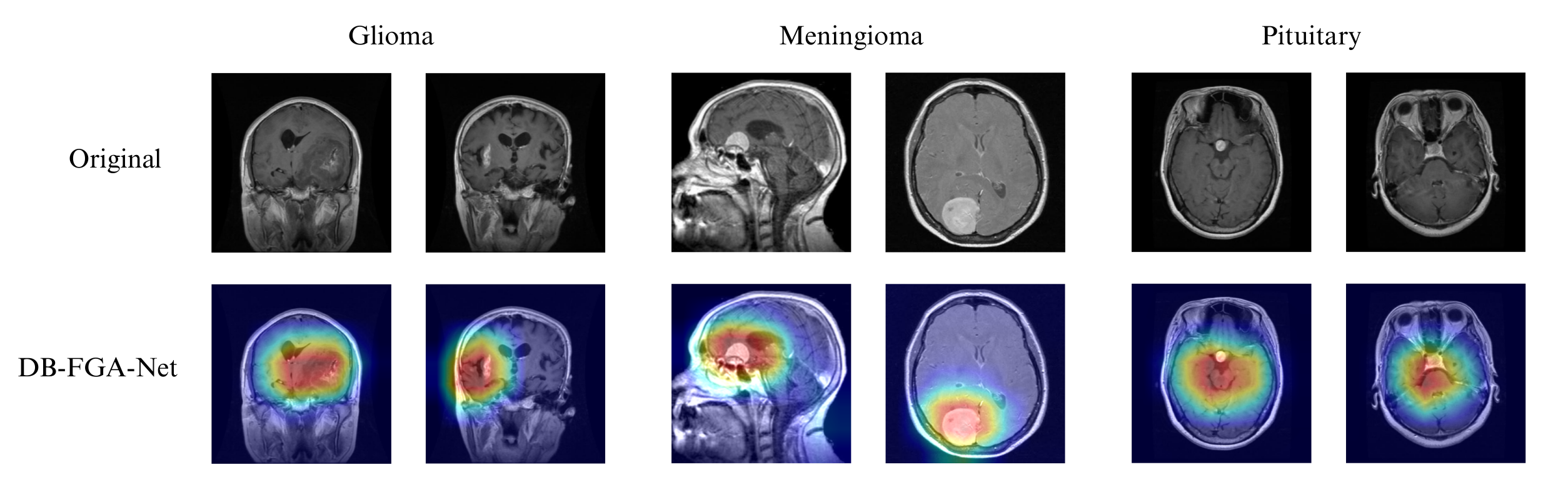}
    \caption{Grad-CAM visualization and analysis of tumor classes for the proposed DB-FGA-Net model}
    \label{fig:7}
\end{figure}

High performance alone is insufficient in medical AI interpretability is equally critical. To this end, Gradient-weighted Class Activation Mapping (Grad-CAM)~\cite{selvaraju2017grad} was used to highlight image regions most influential in the model’s decisions. The importance of interpretability lies in ensuring that models are not exploiting spurious correlations (such as background artifacts), but are instead focusing on medically relevant tumor regions.  
\\
Figures~\ref{fig:6} and~\ref{fig:7} show Grad-CAM heatmaps for CBAM-integrated models and the proposed DB-FGA-Net. In almost every case, FGA-based models localized tumor regions more precisely, especially in challenging meningioma cases. For example, while CBAM sometimes activated irrelevant areas, FGA consistently highlighted tumor boundaries and cores, supporting the claim that frequency gating sharpens pathological features.  
\\
From a clinical perspective, such interpretability builds trust among radiologists. When a model not only predicts tumor presence but also shows a heatmap aligning with actual tumor contours, it increases the likelihood of adoption in diagnostic workflows. By integrating Grad-CAM analysis, we demonstrate that DB-FGA-Net is not a “black box,” but a model whose decision-making process can be verified and validated.  

\subsection{Comparative Performance Analysis}

The proposed DB-FGA-Net framework demonstrates improved performance compared to selected baseline models in brain tumor classification, particularly in multi-class settings on the 7K-DS dataset, achieving an accuracy of 99.24\%, precision of 99.24\%, recall of 99.24\%, and F1-score of 99.24\% in the 4-class configuration, as well as 98.68\% across metrics for 3-class and 99.85\% for 2-class, without relying on data augmentation. This indicates competitive performance relative to existing methods summarized in Table~\ref{tab:model_comparison}. For instance, on the 7K-DS dataset, the proposed model shows higher or comparable performance to A. Saeed et al.'s Dual-Branch Gated Attention Network (96.87--99.62\% accuracy, 96.42--99.06\% precision, 96.32--99.05\% recall, 96.73--99.62\% F1-score)~\cite{GGLA}, which requires GAN and simple augmentation to reach its upper bounds, while our approach achieves comparable or improved results in an augmentation-free manner, highlighting its robustness. Similarly, it demonstrates improved performance compared to N. Sivakumar et al.'s CNN + FedAvg + FedProx (97.19\% accuracy, high precision/recall, 97.18\% F1-score)~\cite{FedAvg} and Anees Tariq et al.'s ViT + EfficientNetV2 Ensemble (96\% across all metrics)~\cite{trans}, both of which employ augmentation techniques like rotation and flipping. Compared to R. Preetha et al.'s Hybrid 3B Net + EfficientNetB2 (97.80\% accuracy for 4-class, high precision/recall/F1-score)~\cite{hybrid3b}, the proposed model provides a 1.51\% accuracy improvement in 4-class and maintains a balanced performance across metrics, despite their use of extensive augmentation on combined datasets. On cross-dataset evaluations involving 3K-DS, our model's 95.77\% accuracy, 96.01\% precision, 95.77\% recall, and 95.75\% F1-score demonstrate competitive and consistent performance compared to H. Alshaari et al.'s EfficientNetB0 w/ Dual Reg. (98\% accuracy, 95\% precision, 98.2\% recall, 95.4\% F1-score)~\cite{deep} and N. M. Hussain Hassan et al.'s Fuzzy Thresholding + DL (98.42--99.42\% accuracy, 98.16--98.65\% precision, 98.14--98.26\% recall, 98.1--98.65\% F1-score)~\cite{fuzzy}, which uses rotation augmentation. Furthermore, against P. Chauhan et al.'s Patch-Based Vision Transformer on Figshare (3-class) (95.8\% accuracy, 95.3\% precision, 93.2\% recall, 92\% F1-score)~\cite{PBviT} and R. D. Prayogo et al.'s ResNet50V2 + MobileNetV2 + DenseNet121 (98.75\% across all metrics)~\cite{hybrid-cnn}, the proposed model provides higher accuracy along with improved interpretability. A key novelty and advantage lies in the emphasis on localization through Grad-CAM, which visualizes tumor-specific regions with high precision (e.g., sharp focus on boundaries in glioma and meningioma), unlike many compared models that lack such interpretability (e.g., PBVit, Hybrid FL, or GGLA-NeXtE2NET). This not only improves diagnostic trust but also addresses clinical needs for explainable AI. Overall, the model demonstrates strong performance without augmentation, making it efficient and potentially generalizable across datasets like 7K-DS and 3K-DS.

\subsection{Graphical User Interface (GUI) Demonstration}

\begin{figure}[H]
    \centering
    % First row
    \subfloat[Glioma]{\includegraphics[width=0.4\textwidth]{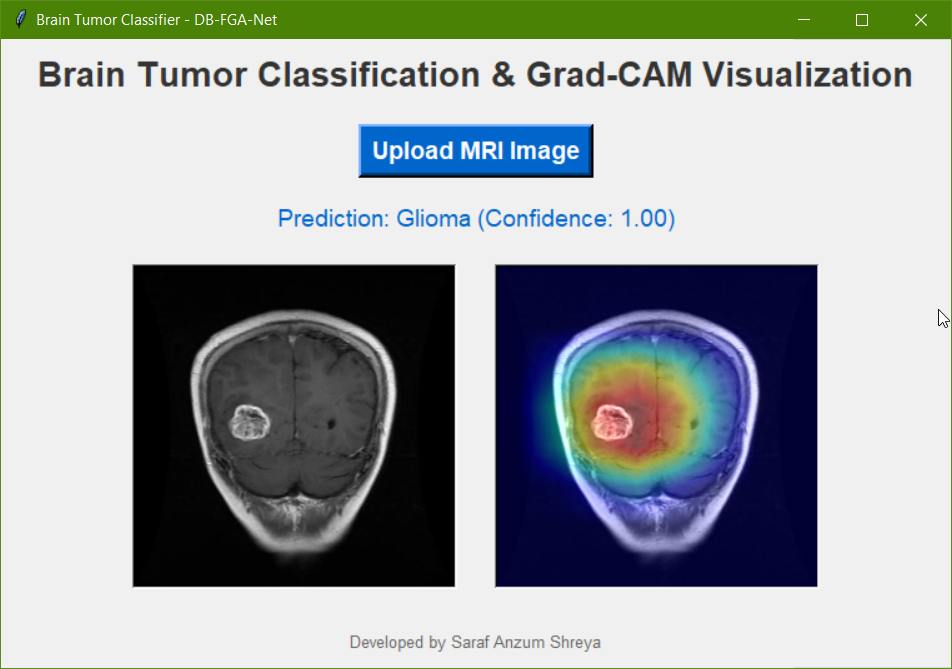}}
    \hfill
    \subfloat[Meningioma]{\includegraphics[width=0.4\textwidth]{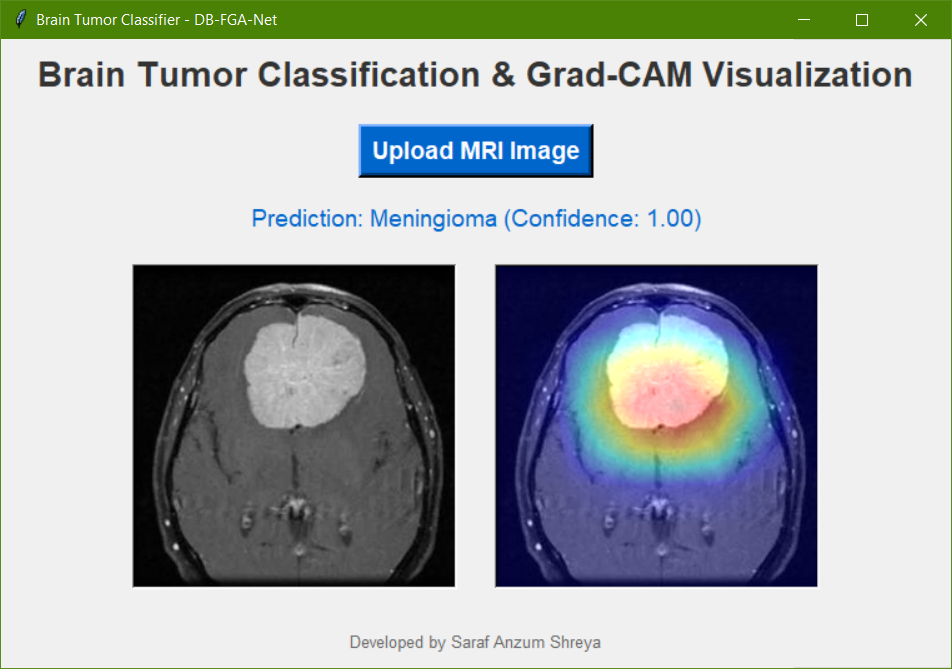}} \\[1ex]
    
    % Second row
    \subfloat[No Tumor]{\includegraphics[width=0.4\textwidth]{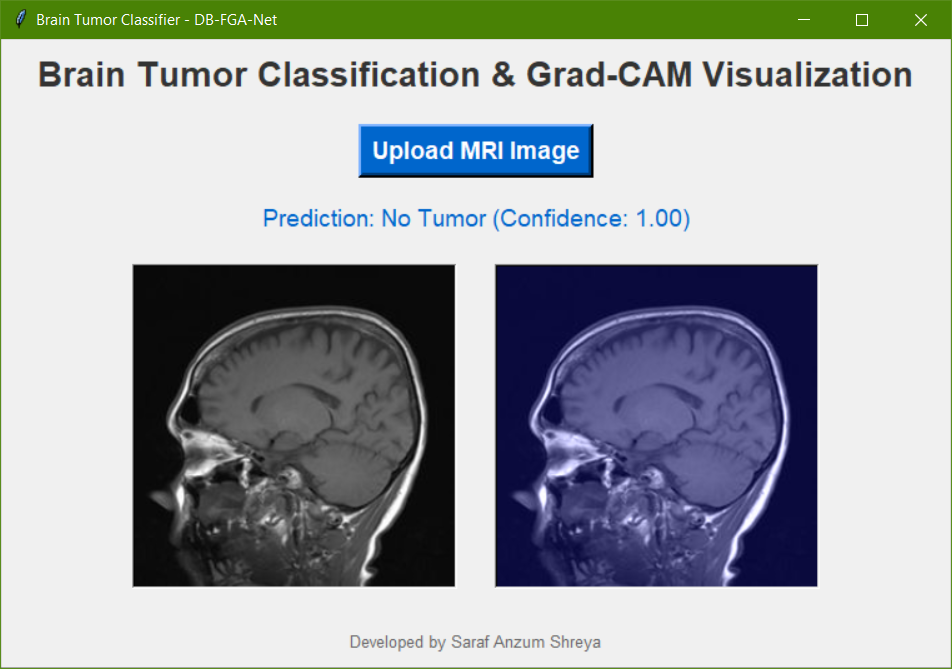}}
    \hfill
    \subfloat[Pituitary]{\includegraphics[width=0.4\textwidth]{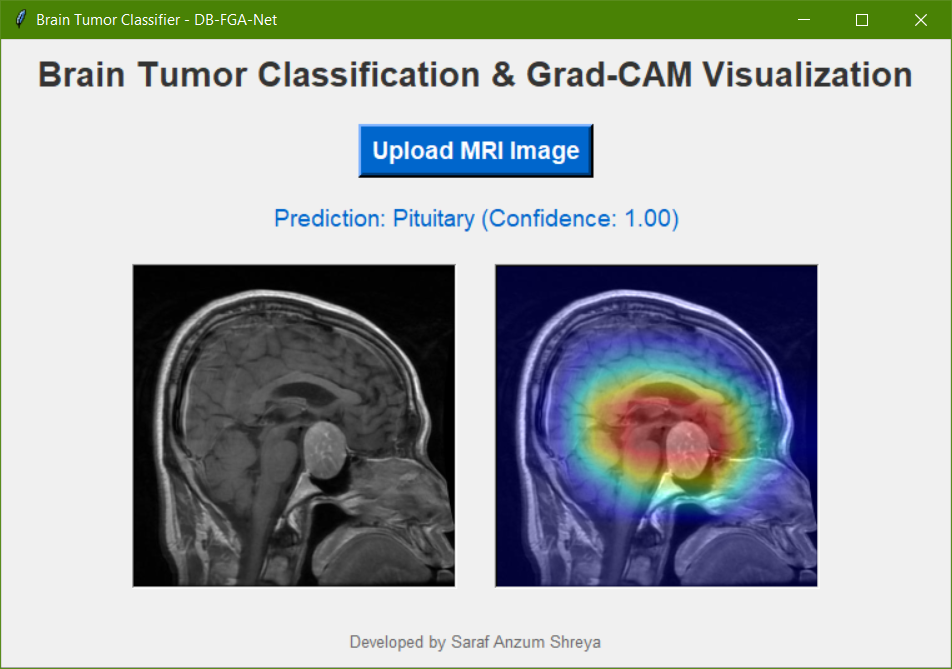}}
    
    \caption{Graphical User Interface (GUI) of DB-FGA-Net. The interface demonstrates predictions and Grad-CAM tumor localization for (a) Glioma, (b) Meningioma, (c) No Tumor, and (d) Pituitary. The left panel shows the original MRI, while the right panel presents the Grad-CAM overlay indicating tumor location.}
    \label{fig:gui_interface}
\end{figure}

To demonstrate the practical usability of DB-FGA-Net, we designed a Graphical User Interface (GUI) using Python's Tkinter library. The GUI allows clinicians and researchers to interactively upload MRI scans, obtain real-time tumor classification results, and visualize the tumor location through Grad-CAM heatmaps. The interface was developed by integrating several Python libraries: TensorFlow for model inference, OpenCV for image processing and Grad-CAM generation, and the Pillow (PIL) library for image handling in the Tkinter environment.
\\
The GUI consists of two main display panels. The left panel shows the original MRI image, while the right panel overlays the Grad-CAM heatmap onto the scan, highlighting the regions most responsible for the model’s prediction. In addition, the classification result with confidence score is displayed below the panels, allowing clinicians to quickly verify both the predicted tumor type and the corresponding anatomical region of interest. This design ensures that the GUI not only provides classification results but also explains the decision by localizing the tumor, thereby enhancing interpretability and clinical trust.
\\
Figure~\ref{fig:gui_interface} presents four representative GUI outputs for each tumor category in the dataset: Glioma, Meningioma, No Tumor, and Pituitary. Each example demonstrates how the system clearly identifies the tumor type and highlights the suspected tumor region, making it a valuable decision-support tool in clinical practice.

\subsection{Strengths of the Proposed Model}
The proposed DB-FGA-Net demonstrates high multi-class accuracy on the primary 7K-DS dataset, achieving 99.24\% accuracy in the 4-class experiment, with correspondingly strong precision, recall, and F1 scores. It maintains excellent performance across different class granularities, achieving 98.68\% in the 3-class setting and 99.85\% in the 2-class setting, highlighting its robustness to varying task difficulty. Beyond the primary dataset, the model generalizes well to an independent dataset (3K-DS) without retraining, preserving strong discriminative ability and demonstrating resilience to domain shifts in acquisition protocols and patient populations.
\\
A key component of the model is the Frequency-Gated Attention (FGA) block, which integrates channel-spatial co-attention with frequency-domain attention and dynamic gating. This design allows the network to capture complementary local textures and global spectral structures, enhancing feature richness without requiring heavy preprocessing. Interpretability is another strength of the model: Grad-CAM visualizations align closely with tumor cores and boundaries in sample cases, providing clinically-relevant saliency maps that can increase radiologist trust. The model also exhibits stable training behavior, supported by 5-fold cross-validation results with low inter-fold variability, underscoring the reliability of its performance across different data splits.
\\
Notably, DB-FGA-Net achieves these results without relying on synthetic data augmentation, simplifying preprocessing and avoiding potential artefacts. In comparison with CBAM-based baselines, DB-FGA-Net consistently outperforms across metrics—for instance, achieving 99.24\% accuracy versus 98.55\% for Xception+CBAM. Grad-CAM visualizations further show tighter and more precise tumor localization compared to CBAM’s diffused activations, highlighting FGA’s enhanced spectral focus and its contribution to improved accuracy and interpretability.
\section{Conclusion}
In this paper, we presented DB-FGA-Net, a dual-backbone network that integrates VGG16 and Xception with a Frequency-Gated Attention (FGA) mechanism for brain tumor classification in MRI images. By combining fine-grained spatial features with complementary frequency-domain cues, the proposed model enhances discriminative power across classes. Experimental evaluation on the 7K-DS benchmark demonstrated that DB-FGA-Net achieves strong and consistent performance across multiple datasets, with an accuracy of 99.24\% in the 4-class setting, 98.68\% in the 3-class setting, and 99.85\% in the 2-class setting. Cross-dataset testing on the independent 3K-DS dataset further confirmed the model’s generalization ability, though some degradation under domain shifts was observed. Grad-CAM visualizations highlighted tumor regions and boundaries, providing interpretability and supporting clinical trust in the model’s predictions. 
\\
To bridge the gap between research and clinical application, we also developed a graphical user interface (GUI) that enables real-time classification and visualization of tumor locations via Grad-CAM overlays. This interface demonstrates the practicality of deploying DB-FGA-Net as a decision-support tool for radiologists and clinicians. 
\\
Future work will focus on addressing domain-shift sensitivity, validating the model on larger multi-center datasets, and optimizing computational efficiency for real-time deployment in clinical workflows. These steps will be critical for ensuring robust, scalable, and reliable clinical translation of the proposed framework.

% \section*{Acknowledgment}
% The preferred spelling of the word ``acknowledgment'' in American English is
% without an ``e'' after the ``g.'' Use the singular heading even if you have
% many acknowledgments. Avoid expressions such as ``One of us (S.B.A.) would
% like to thank $\ldots$ .'' Instead, write ``F. A. Author thanks $\ldots$ .'' In most
% cases, sponsor and financial support acknowledgments are placed in the
% unnumbered footnote on the first page, not here.

\bibliographystyle{ieeetr}

\end{document}